\documentclass[acmtog]{acmart}

\usepackage{booktabs} 
\usepackage{bm}
\usepackage{subcaption}
\usepackage{enumitem}
\usepackage[ruled]{algorithm2e} 
\usepackage{algpseudocode}
\usepackage{tabularx}
\usepackage{hyperref}

\acmJournal{TOG}
\begin{document}

\title{ColorizeDiffusion v2: Enhancing Reference-based Sketch Colorization Through Separating Utilities}

	\author{Dingkun Yan}
	\affiliation{%
		  \institution{Institute of Science Tokyo}
		  \department{School of Computing}
		  \country{Japan}}
	\email{yan@img.cs.titech.ac.jp}
 
	\author{Xinrui Wang}
	\affiliation{%
		  \institution{University of Tokyo}
		  \country{Japan}}
          \email{secret_wang@outlook.com}
  
  	\author{Yusuke Iwasawa}
	\affiliation{%
		  \institution{University of Tokyo}
		  \country{Japan}
        }

	\author{Yutaka Matsuo}
	\affiliation{%
		  \institution{University of Tokyo}
		  \country{Japan}
        }
  
	\author{Suguru Saito}
	\affiliation{%
		  \institution{Institute of Science Tokyo}
		  \department{School of Computing}
		  \country{Japan}}
        \email{suguru@img.cs.titech.ac.jp}
 
	\author{Jiaxian Guo}
	\affiliation{%
		  \institution{University of Tokyo}
		  \country{Japan}}
        \email{jiaxianguo07@gmail.com}
        
\begin{teaserfigure}
        \centering
        \includegraphics[width=0.98\linewidth]{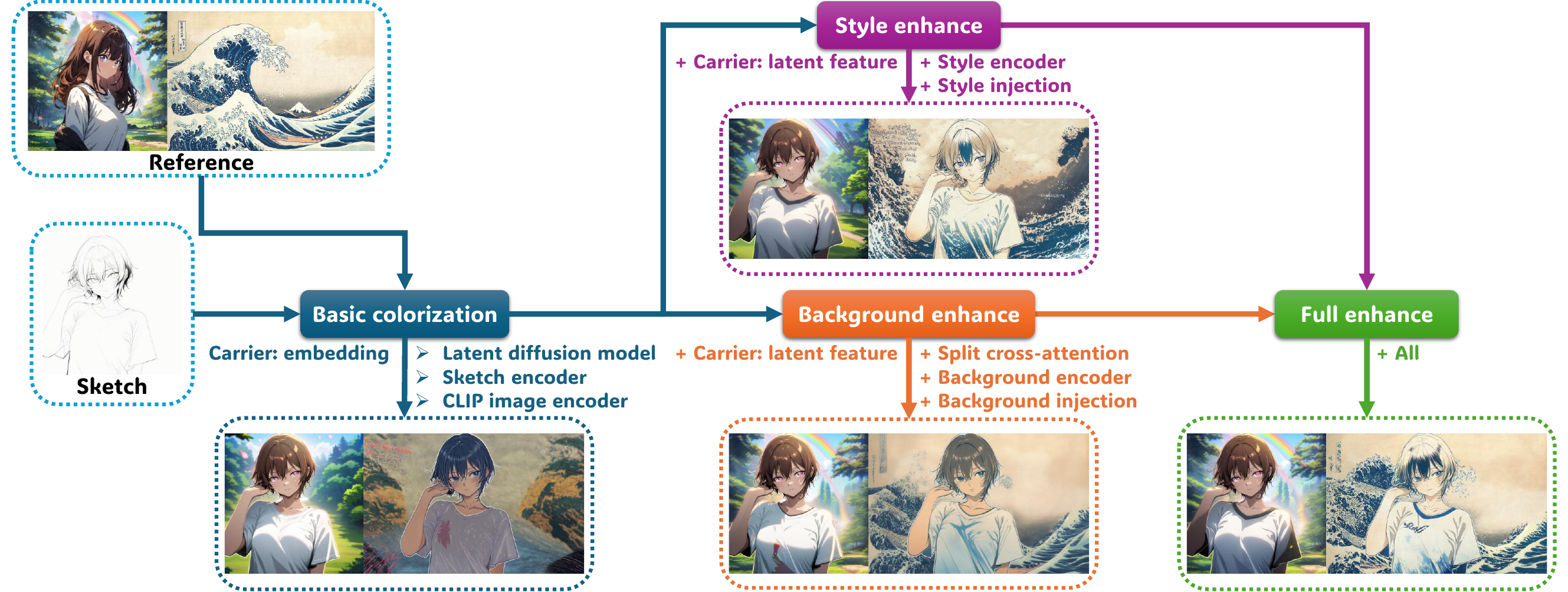}
        \captionof{figure}{We propose a novel reference-based sketch colorization framework that separates the colorization into different aspects and enhances them with respective modules. This system can achieve state-of-the-art anime-style colorization results without requiring spatial correspondence between inputs.}
        \label{teaserfigure}
\end{teaserfigure}

\begin{abstract}
Reference-based sketch colorization methods have garnered significant attention due to their potential applications in the animation production industry. However, most existing methods are trained with image triplets of sketch, reference, and ground truth that are semantically and spatially well-aligned, while real-world references and sketches often exhibit substantial misalignment. This mismatch in data distribution between training and inference leads to overfitting, consequently resulting in spatial artifacts and significant degradation in overall colorization quality, limiting potential applications of current methods for general purposes. To address this limitation, we conduct an in-depth analysis of the \textbf{carrier}, defined as the latent representation facilitating information transfer from reference to sketch. Based on this analysis, we propose a novel workflow that dynamically adapts the carrier to optimize distinct aspects of colorization. Specifically, for spatially misaligned artifacts, we introduce a split cross-attention mechanism with spatial masks, enabling region-specific reference injection within the diffusion process. To mitigate semantic neglect of sketches, we employ dedicated background and style encoders to transfer detailed reference information in the latent feature space, achieving enhanced spatial control and richer detail synthesis. Furthermore, we propose character-mask merging and background bleaching as preprocessing steps to improve foreground-background integration and background generation.
Extensive qualitative and quantitative evaluations, including a user study, demonstrate the superior performance of our proposed method compared to existing approaches. An ablation study further validates the efficacy of each proposed component. We show the colorization workflow with a fully-featured UI and our code are available in \url{https://github.com/tellurion-kanata/colorizeDiffusion}.

\end{abstract}

	\ccsdesc[500]{Applied computing~Fine arts}
	\ccsdesc[500]{Computing methodologies~Computer vision}
	\ccsdesc[300]{Computing methodologies~Image processing}
	
	\keywords{Sketch colorization, Image-guided generation, Latent diffusion model}
    
\maketitle

\section{Introduction}
\label{Introduction}

Animation, with a history tracing back to the advent of celluloid film animation in 1888, has remained a prominent artistic form for centuries. Initially relying on hand-painted frames on paper and celluloid, the production workflow has since transitioned to digital platforms utilizing software such as CLIP Studio and Adobe Animate. Contemporary animation production, however, remains labor-intensive, and increasing market demand is straining studio capacity, posing significant challenges to the industry.

Within current animation creation workflows, sketch colorization represents a particularly labor-intensive process, occupying a substantial portion of studio personnel. Consequently, machine learning techniques have been explored to automate this task and alleviate manual effort. Early attempts utilizing Generative Adversarial Networks (GANs) \cite{IsolaZZE17,zhang2017style,ZhangLW0L18,zouSA2019sketchcolorization,yan-cgf} yielded suboptimal colorization results due to limitations in their generative capacity. 

\begin{figure}[t]
    \centering
    \includegraphics[width=\linewidth]{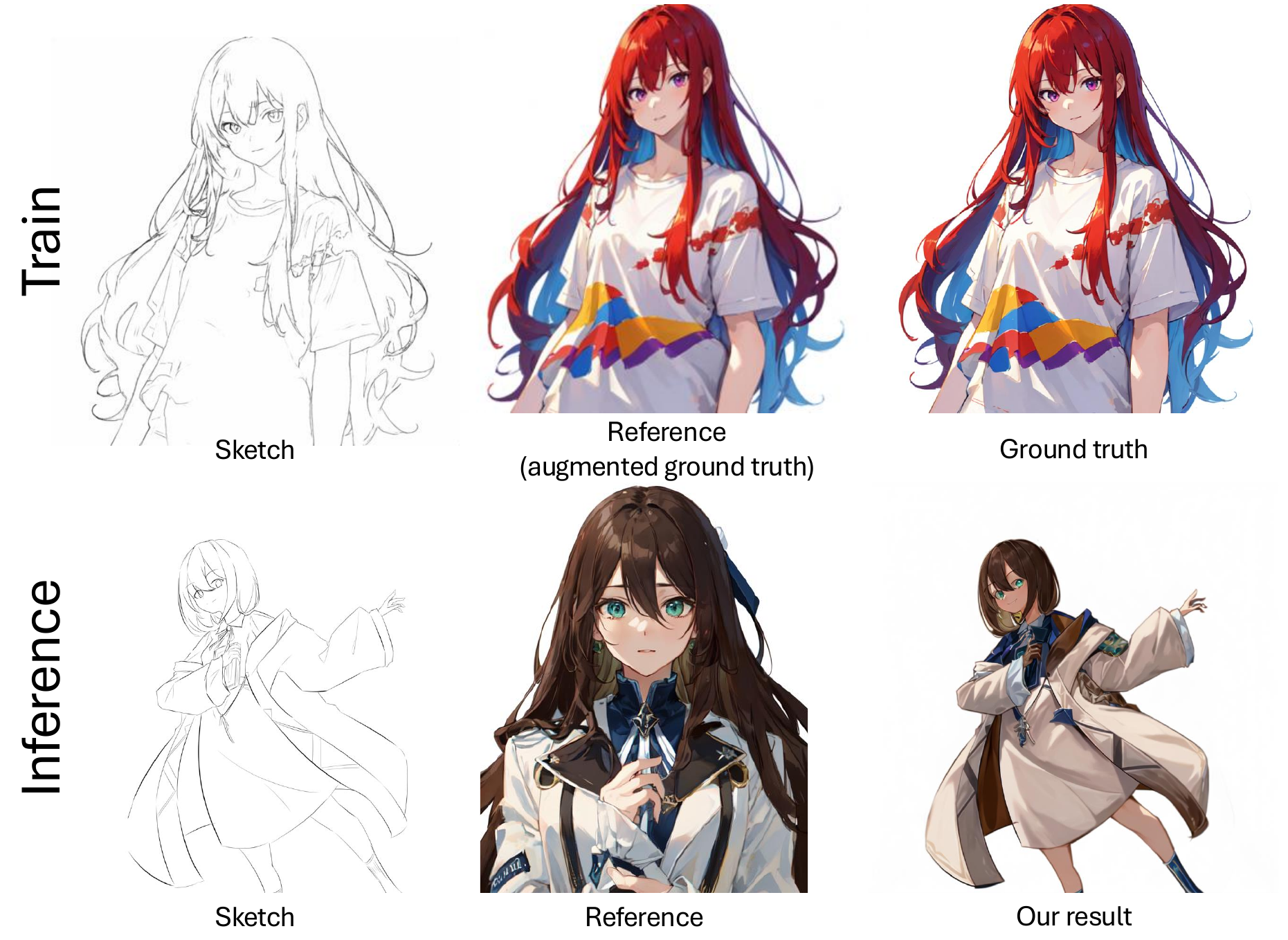}
    \vspace{-1.5em}
    \caption{An example of distribution shift in training and inference data. Existing reference-based methods typically derive references from ground truth with deformation due to the data limitation, the augmented references are still highly semantically related to the sketches. During inference, however, the reference images could be non-relevant to the sketches. This distribution gap between train and inference data results in overfitting, severely detereiorate the colorization quality.}
    \label{training-sample}
\vspace{-1.5em}
\end{figure}

Recently, diffusion models have demonstrated notable success in image synthesis and have subsequently been applied to sketch colorization. Reference-based methods have garnered particular attention due to their potential to be seamlessly integrated into existing production pipelines. Due to the lack of datasets and the difficulty of collecting large amounts of semantically aligned but spatially mismatched image triplets, existing image-guided approaches typically derive training sketches and references by extracting and deforming ground truth color images, respectively, as illustrated in Figure \ref{training-sample}.

\begin{figure}[t]
    \centering
    \includegraphics[width=0.98\linewidth]{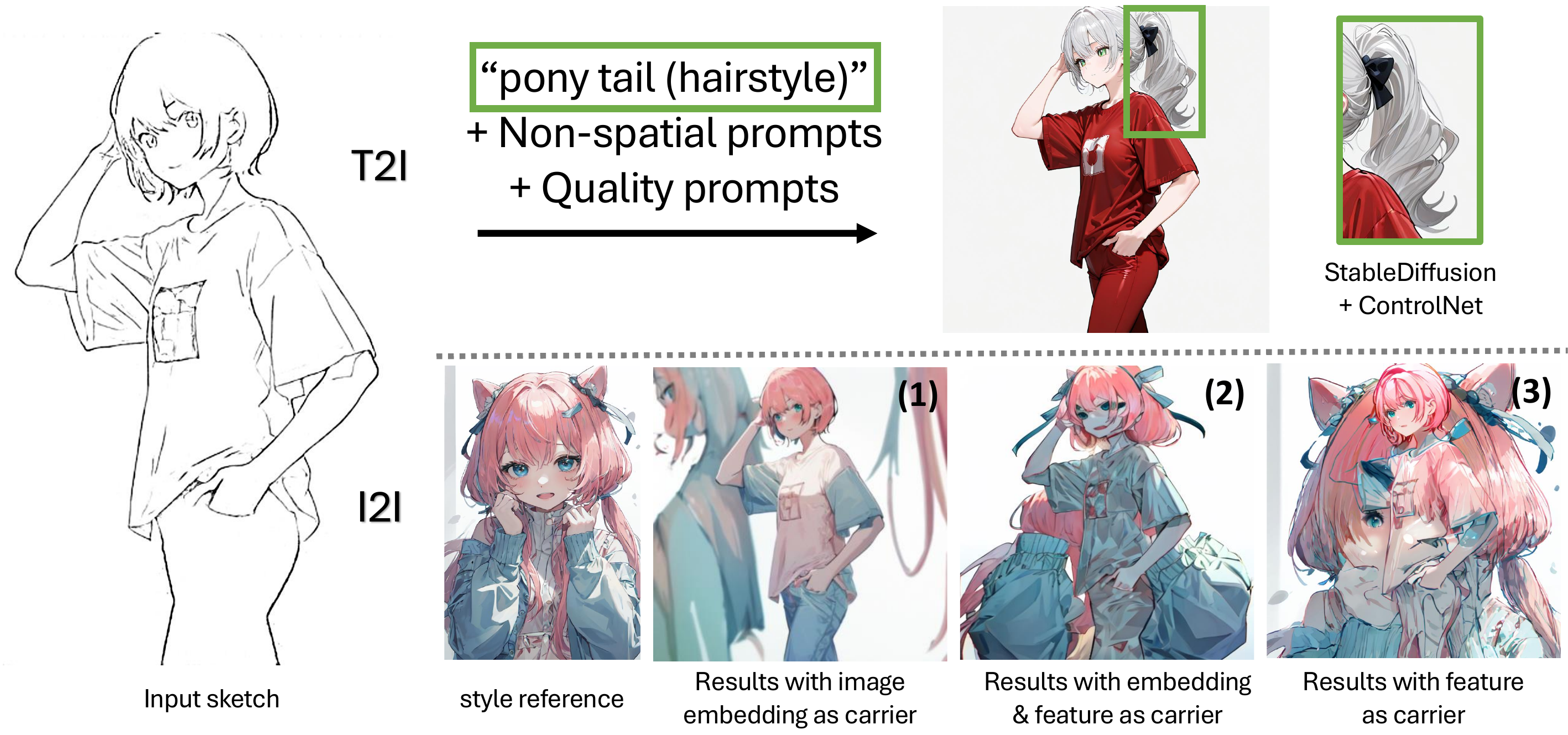}
    \vspace{-0.5em}
    \caption{Spatial entanglement represented in different ways. The T2I method mistakenly changes hairstyle in the results. For the I2I methods, spatial entanglements are influenced by the carrier used for the color reference injection. Carriers containing less detail and higher-level semantic information tend to produce results with fewer artifacts and blurry textures, and vice versa. Source of results: 1. our vanilla mode; 2. fine-tuned IP-Adapter; 3. jointly-trained reference net.}
    \label{entanglement}
\vspace{-1.5em}
\end{figure}

However, this training paradigm introduces a distribution shift during inference, as reference images and sketches are no longer derived from the same source \cite{yan2024colorizediffusion}. This discrepancy between training and inference data distributions presents a significant challenge to reference-based sketch colorization, as overfitting to the training distribution leads to artifacts and severe deterioration in image quality for inputs with misaligned semantics or structures. While user inputs are mostly unpaired images in practice, it significantly limits the application of image-guided colorization methods compared to their text-guided counterparts. Specifically, images in current digital illustration and animation workflows typically feature characters in the foreground and scenery in the background, and such conflicts often manifest as severe distortions or inaccuracies in the generated semantic structure.
We further observe that these negative impacts are influenced by the carrier used to transfer reference information. Carriers containing less detail and higher-level semantic information tend to produce results with fewer artifacts and blurry textures, while carriers with richer details and lower-level semantic information yield results with better textures but more pronounced artifacts. Termed "spatial entanglement," these bad cases are illustrated with corresponding carriers in Figure \ref{entanglement}.

To eliminate artifacts such as extra characters or body parts in the background, we employ spatial masks to segment foreground and background regions within both the sketch and reference images. This enables the network to process these regions independently through a novel split cross-attention mechanism, where embeddings of the segmented regions are injected separately into the diffusion backbone. Subsequently, we introduce a background encoder and a style encoder after pre-training the core colorization network. These encoders enhance spatial control and facilitate the transfer of fine-grained details from the reference image while preserving the network's understanding of the sketch semantics. This is achieved by training the encoders with the diffusion backbone frozen, preventing the network from compromising its initial understanding of the sketch during the encoder training phase. Finally, we employ a step-by-step training strategy to optimize each module, and propose character-mask merging and background bleaching techniques to further refine foreground-background blending and background synthesis, effectively mitigating specific types of artifacts.

We collect and process a large-scale dataset, comprising 6.5 million training images and 50,000 validation images, encompassing diverse scenarios and use cases. Our experimental evaluation includes a comprehensive ablation study demonstrating the contribution of each component in mitigating spatial entanglement and artifacts in practical applications. Qualitative analysis shows that our method generates high-quality results that faithfully transfer the color distribution from reference images while avoiding spatial entanglement. Quantitative comparisons against existing methods with established metrics and benchmarks further validate the superiority of our approach. Finally, user studies provide subjective evidence of user preference for our method over existing alternatives.

In summary, our contributions are threefold: (1) We provide a detailed analysis of current reference-based sketch image colorization methods and identify the underlying causes and manifestations of artifacts. (2) We propose a novel colorization framework with components designed to address specific artifact types, resulting in effective mitigation and high-quality colorization without requiring well-aligned input pairs. (3) Extensive experiments demonstrate the superiority of our method over existing approaches through qualitative and quantitative comparisons, as well as positive user feedback in perceptive user study.

\begin{figure*}[t]
    \centering
    \includegraphics[width=1\linewidth]{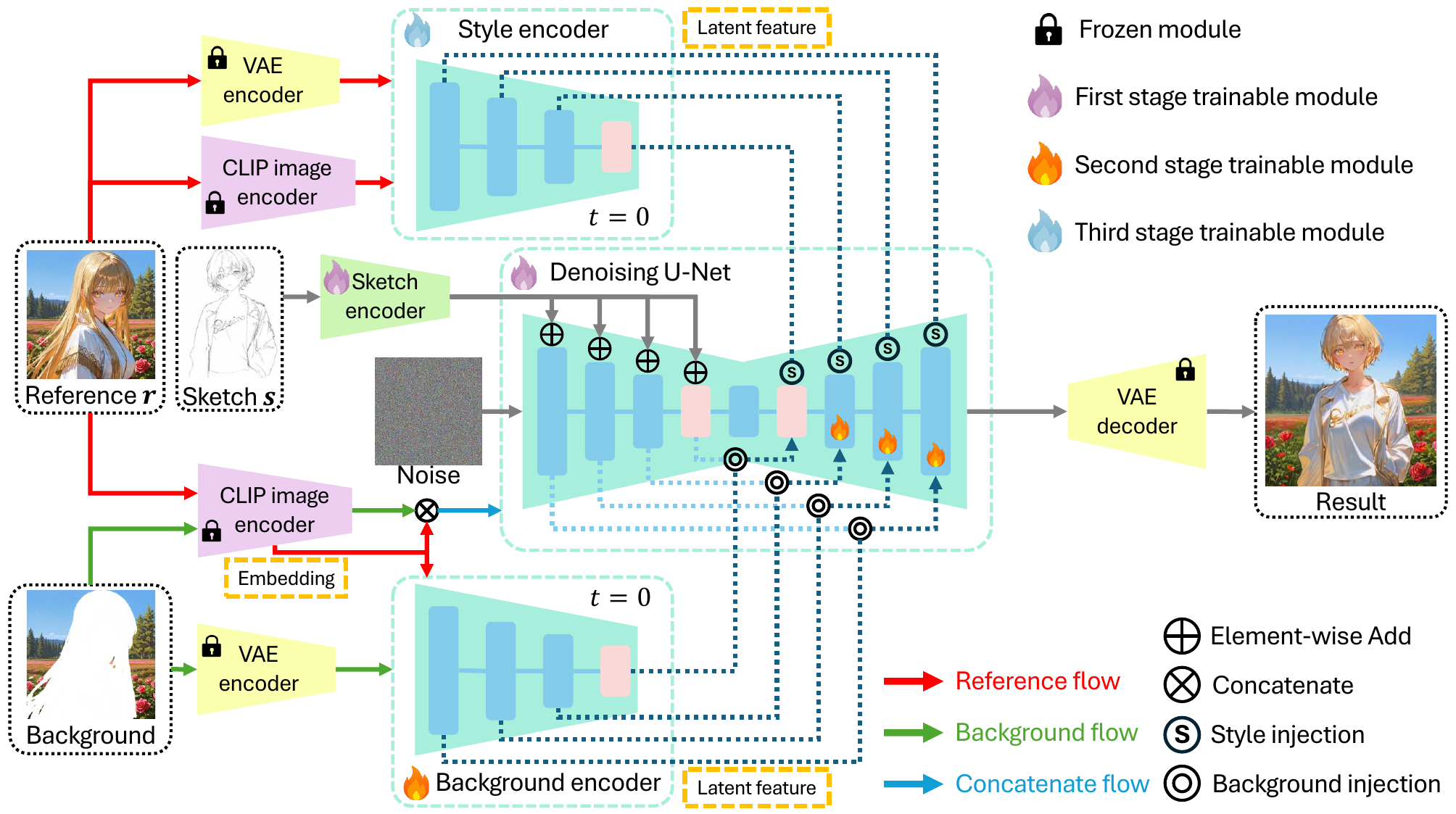}
    \caption{Illustration of the proposed framework with extracted carriers labeled by orange rectangles next to their corresponding encoder. We use masks to separate reference images into foreground and background for CLIP Image encoder $\phi$ to extract respective embeddings. Their concatenation is K and V inputs for split cross-attention in the denoising U-Net, and background embeddings are KV inputs for the background U-Net encoder. Note that we directly use ground truth color images as references in all the training mentioned in this paper. Sketch and mask images are generated from original color images using existing extraction methods \cite{sketchKeras,anime-segmentation}.}
    \label{framework}
\end{figure*}
\begin{figure}[t]
    \centering
    \includegraphics[width=1\linewidth]{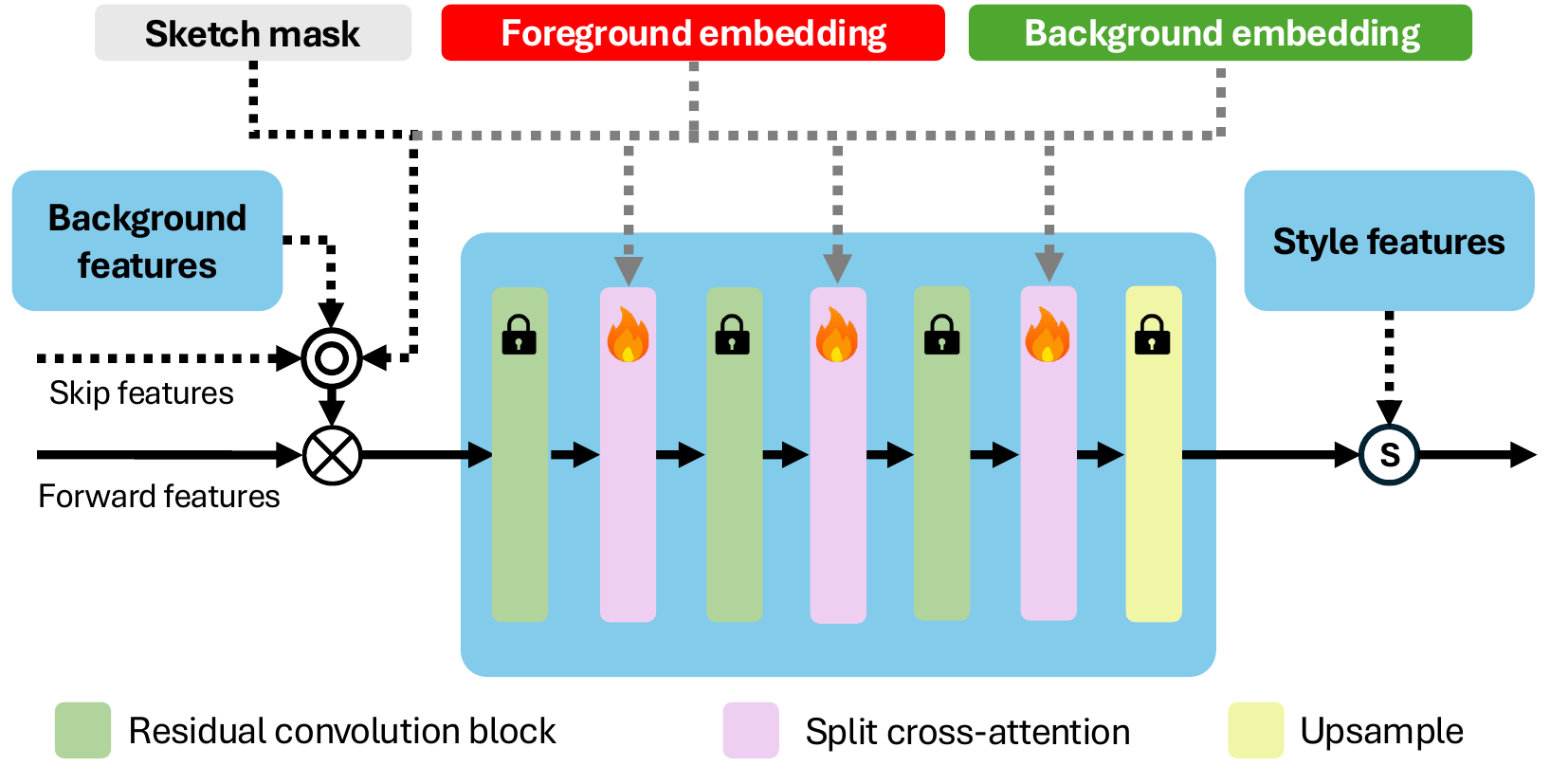}
    \caption{Illustration of a decoder block in the denoising U-Net after merging the split cross-attention, the background encoder, and the style encoder. Note that only the LoRA weights in the split cross-attention are optimized in the proposed training.}
    \label{decoder-block}
\end{figure}
\section{Related Work}
\subsection{Latent Diffusion Models}

Diffusion Probabilistic Models \cite{HoJA20,0011SKKEP21} are a class of latent variable models inspired by nonequilibrium thermodynamics \cite{Sohl-DicksteinW15} and have achieved great success in image synthesis and editing. Compared to Generative Adversarial Networks (GANs)\cite{GoodfellowPMXWOCB14,KarrasLA19,KarrasLAHLA20,ChoiCKH0C18,ChoiUYH20}, Diffusion Models excel at generating high-quality images with various contexts and stronger control over different conditional guidance. However, the autoregressive denoising process of diffusion models, typically computed with a U-Net \cite{RonnebergerFB15} or a Diffusion Transformer (DiT) \cite{DiT,pixart}, incurs substantial computational costs. 

To reduce this cost, Rombach et al. proposed Stable Diffusion (SD) \cite{RombachBLEO22,sdxl}, a class of Latent Diffusion Models (LDMs) that performs diffusion/denoising process inside a perceptually compressed latent space using a pair of pre-trained Variational Autoencoder (VAE). Concurrently, several studies on accelerating the denoising process have demonstrated their effectiveness \cite{SongME21,0011SKKEP21,0011ZB0L022,abs-2211-01095}. In this paper, we adopt SD as our neural backbone, utilize the DPM++ solver \cite{abs-2211-01095,0011SKKEP21,KarrasAAL22} as the default sampler, and employ classifier-free guidance \cite{DhariwalN21,abs-2207-12598} to strengthen the transfer performance.

\subsection{Image Prompted Diffusion Models}

Existing deep generative methods have achieved notable progress in text-guided generation \cite{RombachBLEO22,sdxl,DiT,pixart}. However, many tasks require more detailed guidance than texts can provide for better control of the generated content, including image-to-image translation \cite{KwonY23}, style transfer \cite{instantstyle, zhang2023inversion}, colorization \cite{animediffusion,yan2024colorizediffusion} and image composition \cite{zhang2023controlcom,kim2023reference}. In these cases, images are used as prompts to provide reference information. The reference information extracted from prompt images varies from task to task: style transfer prefers textures and colors from reference images, image composition focuses more on object-related information, and sketch colorization requires all above.
 
Two primary carriers are commonly employed to transfer reference information to diffusion models: image embeddings extracted by pre-trained vision encoders \cite{radford2021learning,openclip,openclip-2,ZhangLW0L18,yan-cgf,yan2024colorizediffusion,t2i-adapter} and latent features directly encoded from reference images by jointly trained modules \cite{animediffusion,hu2023animateanyone,lvcd,tooncraft,li2022eliminating,ip-adapter}. However, these carriers can introduce mismatched structural and semantic information, particularly when the input sketch and reference image are poorly aligned, leading to degraded generation quality. In the context of sketch colorization, these carriers unavoidably provide conflicting spatial cues during inference, resulting in unacceptable artifacts as depicted in Figure \ref{entanglement}.

\begin{table*}[t]
    \centering
    \caption{Characteristics of reference carriers regarding their information and negative impacts caused by the spatial entanglement. Compared to embedding, entanglement caused by the feature-level carrier is harder to eliminate with existing data and training schemes \cite{yan2024colorizediffusion,lvcd,tooncraft,meng2024anidoc}. The keynote of this paper is to add the feature-level transfer to a well-trained embedding-guided model by separating utilities and training respective adapters step-by-step.}
    \begin{tabular}{|c|p{4cm}|p{6cm}|p{4cm}|}
        \hline 
        Carrier & \centering Common sources & \centering Semantic level & \centering\arraybackslash Entanglement phenomenon \\
        \hline
        Image embeddings & Visual encoder pre-trained for language-related tasks & High-level visual features that could be captioned by natural language & Blurry textures and details; change of sketch semantics \\
        \hline
        Latent features & Trainable modules between reference inputs and diffusion backbone & Lowest-level visual features the module can extract, unable to precisely describe using natural language & Ignoring sketch semantics whose counterparts are from reference during training \\
        \hline
    \end{tabular}
    \label{reference-comp}
\end{table*}
\subsection{Sketch Colorization}

Sketch colorization has been an active research area in computer vision. While early approaches relied on interactive optimization \cite{SykoraDC09}, deep learning has emerged as the dominant paradigm, enabling the synthesis of high-quality, high-resolution color images \cite{ZhangLW0L18, KimJPY19,li2022eliminating,animediffusion,yan2024colorizediffusion}. Based on the guiding modality, existing methods can be broadly categorized into three groups: text-prompted \cite{KimJPY19,yan-cgf,controlnet-iccv}, user-guided \cite{ZhangLW0L18,s2pv5}, and reference-based \cite{li2022eliminating,yan-cgf,animediffusion}. User-guided methods offer fine-grained control through direct user input (e.g., spots, sprays), but their reliance on manual intervention hinders their applicability in automated pipelines. Text-prompted methods, driven by advancements in text-to-image diffusion models, have gained significant traction, yet it is challenging to precisely control colors, textures, and styles using text prompts. 

Image-referenced methods have also benefited from advancements in diffusion models and image control techniques \cite{controlnet-iccv,controllllite,controlnet-v11,t2i-adapter,ip-adapter}. However, a critical challenge remains: effectively addressing the spatial and semantic conflicts between diverse reference images with the often sparse and abstract nature of sketches during inference. Existing methods typically require extracted sketches with highly-matched references as input pairs to achieve satisfying results \cite{li2022eliminating,yan-cgf,animediffusion,tooncraft,meng2024anidoc,lvcd}, limiting their generalizability and potential applications. While ColorizeDiffusion \cite{yan2024colorizediffusion} demonstrated significant progress in colorization quality, it still grapples with spatial entanglement, as visualized in Figure \ref{entanglement}, and struggles to accurately transfer fine-grained details. In this paper, we address these limitations by introducing a novel step-by-step training strategy within a refined colorization framework designed to explicitly mitigate spatial entanglement.

\section{Architecture}

In section \ref{Introduction} and figure \ref{training-sample}, we show that the distribution shift between data used for training and inference of reference-based sketch colorization methods leads to bad cases and artifacts called spatial entanglement. The common practice of existing methods is to employ a feature extraction network to encode the reference image into a certain carrier and inject it into the diffusion backbone. Illustrated in figure \ref{entanglement}, the characteristics of spatial entanglement are influenced by the carrier used for the color reference injection. Especially, using image embeddings that contain high-level semantic information and fewer details as carrier leads to fewer artifacts but blurry textures in the results; on the contrary, using latent features with low-level semantic information and richer detailed information as carrier results in severe artifacts but finer textures in the results. To eliminate spatial entanglement and improve the colorization quality, we propose a colorization framework that employs both image embeddings and latent features as colorization carriers and achieves artifacts-free high-quality sketch colorization.

The framework consists of a diffusion backbone, a variational autoencoder (VAE), a sketch encoder, a CLIP image encoder, a style encoder, and a background encoder. It leverages a sketch image $\bm{X_s} \in \mathbb{R}^{w_s \times h_s \times 1}$, a reference image $\bm{X_r} \in \mathbb{R}^{w_r \times h_r \times c}$, a sketch mask $\bm{X_{sm}} \in \mathbb{R}^{w_s \times h_s \times 1}$ and a reference mask $\bm{X_{rm}} \in \mathbb{R}^{w_r \times h_r \times 1}$ as inputs, and returns the colorized result $\bm{Y} \in \mathbb{R}^{w_s \times h_s \times c}$, with $w$, $h$ and $c$ representing the width, height and channel of the images.

\subsection{Reference extraction and carriers}

In this paper, we note the intermediate representations used to transfer information from the reference image to the diffusion backbone as \textbf{carrier}. Existing sketch colorization methods typically employ a feature extraction network to derive carriers from reference images, which are subsequently injected into a diffusion backbone. As illustrated in figure \ref{entanglement}, both image embeddings \cite{ip-adapter,yan2024colorizediffusion} and latent features \cite{lvcd,tooncraft} have been utilized as the carriers, and the characteristics of spatial entanglement are influenced by the carrier used for the color reference injection. 
The image embeddings are typically the outputs of the last layers of pre-trained vision encoder for language-related tasks, such as CLIP image encoder for image caption or Transformers for multi-label classification. They encapsulate high-level semantic information but possess limited detail as a carrier, consequently tend to yield results with fewer artifacts but exhibit blurry textures. The latent features are usually extracted from reference images by trainable modules, such as reference net in \cite{hu2023animateanyone} or the U-Net encoder itself in \cite{li2022eliminating,animediffusion}. Opposite to image embeddings, they contain low-level semantic information and richer detail as carrier, which results in much easier overfitting and more pronounced artifacts but finer textures in the results. 

In this work, we leverage the advantages of both carrier types while mitigating their respective limitations to achieve more accurate and higher-quality colorization. To extract image embeddings, we make use of the image encoder from OpenCLIP-H \cite{RadfordKHRGASAM21,openclip,openclip-2,schuhmann2022laionb}. The pre-trained ViT-based image encoder network extracts 2 kinds of image embeddings: the CLS embeddings $\bm{E_{cls}} \in \mathbb{R}^{bs \times 1 \times 1024}$ and the local embeddings $\bm{E_{local}} \in \mathbb{R}^{bs \times 256 \times 1024}$, where $bs$ represent the batch size. Previous image-guided diffusion models \cite{ip-adapter, instantstyle} commonly utilize CLS embeddings as color or style references, which are connected to text-level notions and projected to CLIP embedding space for image-text contrastive learning, with spatial information compressed. ColorizeDiffusion \cite{yan2024colorizediffusion}, on the other hand, reveals that local embeddings also express text-level semantics and, meanwhile, express more details regarding textures, strokes, and styles, enabling the network to generate better reference-based results with finer details. Consequently, the proposed method follows \cite{yan2024colorizediffusion} to adopt local embeddings as a colorization carrier.

Nevertheless, the inherent lack of low-level detailed reference information in image embeddings can lead to blurry textures in the colorized output. For sketches depicting isolated figures or featuring simple background lines, the information encoded within image embeddings may be insufficient to reconstruct a meaningful and visually compelling background. To address this limitation, we incorporate a style encoder and a background encoder, trained to extract latent features that are also injected into the diffusion backbone as colorization references. We will introduce the mechanism that separately processes foreground and background with corresponding carriers to prevent spatial entanglements in section \ref{split_cross_attention} and the detail of the two encoders in section \ref{background_encoder} and section \ref{style_encoder}.

\begin{figure}[t]
    \centering
    \includegraphics[width=1\linewidth]{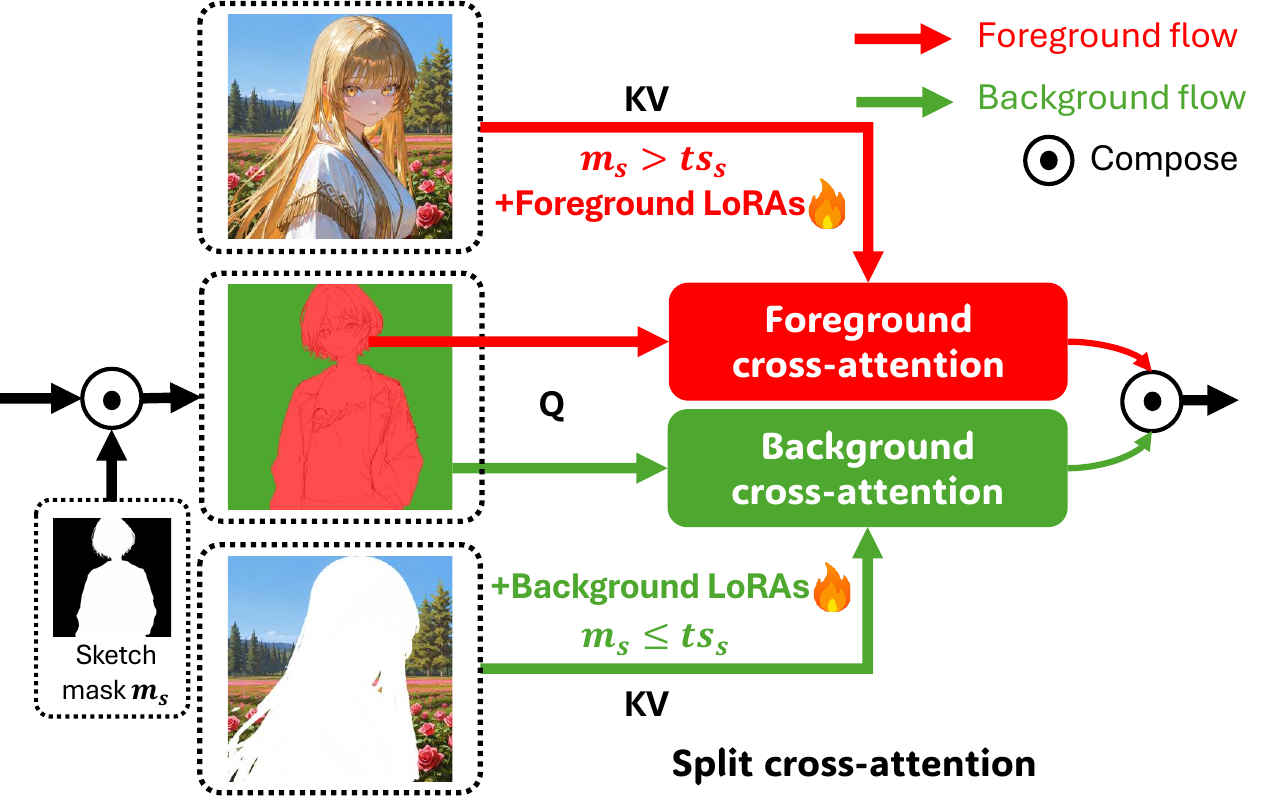}
    \caption{We replace all cross-attention in the denoising U-Net with the proposed split cross-attention, where foreground and background regions receive respective embeddings during one forward pass. Additionally, we add learnable foreground/background LoRA weights in the split cross-attention in the decoder part of U-Net. Corresponding formulation is given in Eq. \ref{split-attention-eq}.} 
    \label{split-crossattn-arch}
\end{figure}

\subsection{Separate process of foreground and background}
\label{split_cross_attention}

In existing reference-based sketch colorization methods, spatial entanglements mainly represent as foreground exceeding boundaries and appearing in the background, such as extra characters or body parts. An efficient way to eliminate the phenomena is to separately process the foreground and background with independent modules. Following the common use case of digital illustration and animation production, we define sketch-guided regions as the foreground, which are mostly characters in practice, and the rest areas of the images as the background. Animation segmentation methods \cite{anime-segmentation,isnet-eccv2022} are then employed to obtain spatial masks for sketches and reference images.

Based on this foreground/background separation, we further implement a novel split cross-attention mechanism. Functioning as cross-attention layers in the diffusion backbone, it separately processes foreground and background regions with different parameters in a single forward pass as visualized in Figure \ref{split-crossattn-arch}. 
Given foreground mask $\bm{m}_{s}$ and $\bm{m}_{r}$ of sketch and reference images, regions in respective images with mask values larger than user-defined thresholds $ts_{s}$ and $ts_{r}$ are considered as foreground, otherwise background. We define query inputs (forward features) as $\bm{z}$, key and value inputs (reference embeddings) as $\bm{e}$, $\bm{e}_{bg}$, in the following sections, where the index $bg$ indicate background. Specifically, $\bm{e}$ denotes the reference embeddings extracted from the whole reference image $\bm{r}$, formulated as $\bm{e}=\phi(\bm{r})$; and $\bm{e}_{bg}=\phi(\bm{r}_{bg})$, where $\bm{r}_{bg}$ is the background region of the reference image. During training, the proposed split cross-attention can be formulated as follows:

\begin{equation}
\begin{small}
    \bm{z} = \begin{cases}
    \mathbf{Softmax}(\frac{(\hat{W}^{fg}_{q}\bm{z})\cdot(\hat{W}^{fg}_{k}\bm{e})}{d})(\hat{W}^{fg}_{v}\bm{e}) & \text{if $\bm{m_{s}} > ts_{s}$}\\
    \mathbf{Softmax}(\frac{(\hat{W}^{bg}_{q}\bm{z})\cdot(\hat{W}^{bg}_{k}\bm{e}_{bg})}{d})(\hat{W}^{bg}_{v}\bm{e}_{bg}) & \text{if $\bm{m_{s}}\leq ts_{s}$}
    \end{cases}
    \label{split-attention-eq}
\end{small}
\end{equation}

where $\hat{W}_{t} = W_{t} + W^{i}_{t}$, and $W_{t}$ represents the pre-trained weights. $W^{i}_{t}$ denotes the learnable foreground/background LoRA weights with a rank of 4, which are jointly trained with the background encoder.

\begin{figure}[t]
    \centering
    \includegraphics[width=1\linewidth]{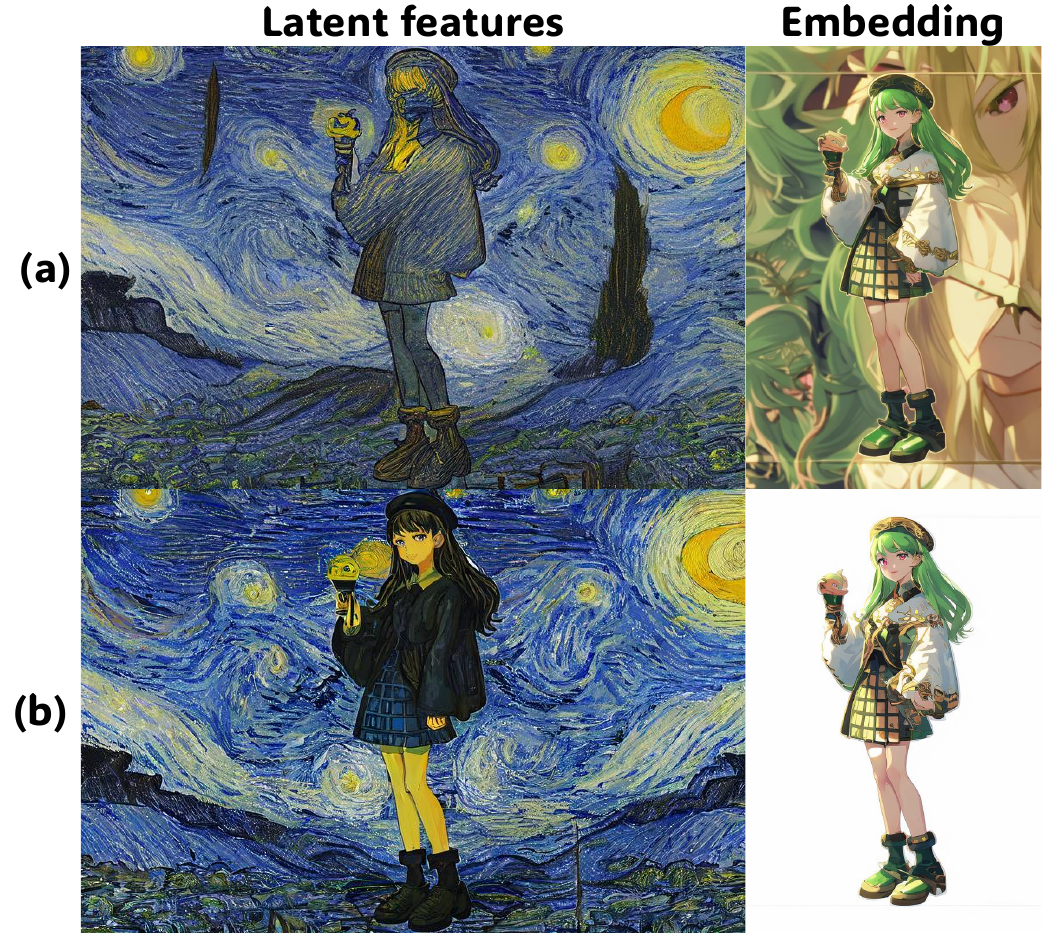}
    \caption{Comparison of representative spatial entanglement caused by different reference carriers: (a) results with entanglement; (b) results with entanglement eliminated by the proposed framework. Networks fully guided by latent features ignore the segmentation of sketches, and those guided by embeddings generate duplicated contents outside sketches, as in this figure, or change the semantics that should be determined by sketches, as visualized in Figure \ref{entanglement}.}
    \label{carrier-comp}
\end{figure}

\subsection{Background encoder and reference injection}
\label{background_encoder}

To synthesize colorization results with enhanced background textures and semantics, particularly for sketches with sparse or absent background lines, we introduce a dedicated background encoder to extract detailed content references. Drawing inspiration from ControlNet \cite{t2i-adapter,controlnet-iccv}, we consider the generated image can be considered the decoded result of the overlapping region of two conditional latent distributions $p(z|r)$ and $p(z|s)$ assuming the denoising backbone is fixed during the training of the additional encoder. Therefore, we adopt a separate U-Net encoder to introduce an independent conditional distribution $p^{*}(z|r)$ based on the reference image. This background encoder only forwards one time during the whole denoising process, with its timestep input set as $t=0$. 

\begin{figure}[t]
    \centering
    \includegraphics[width=1\linewidth]{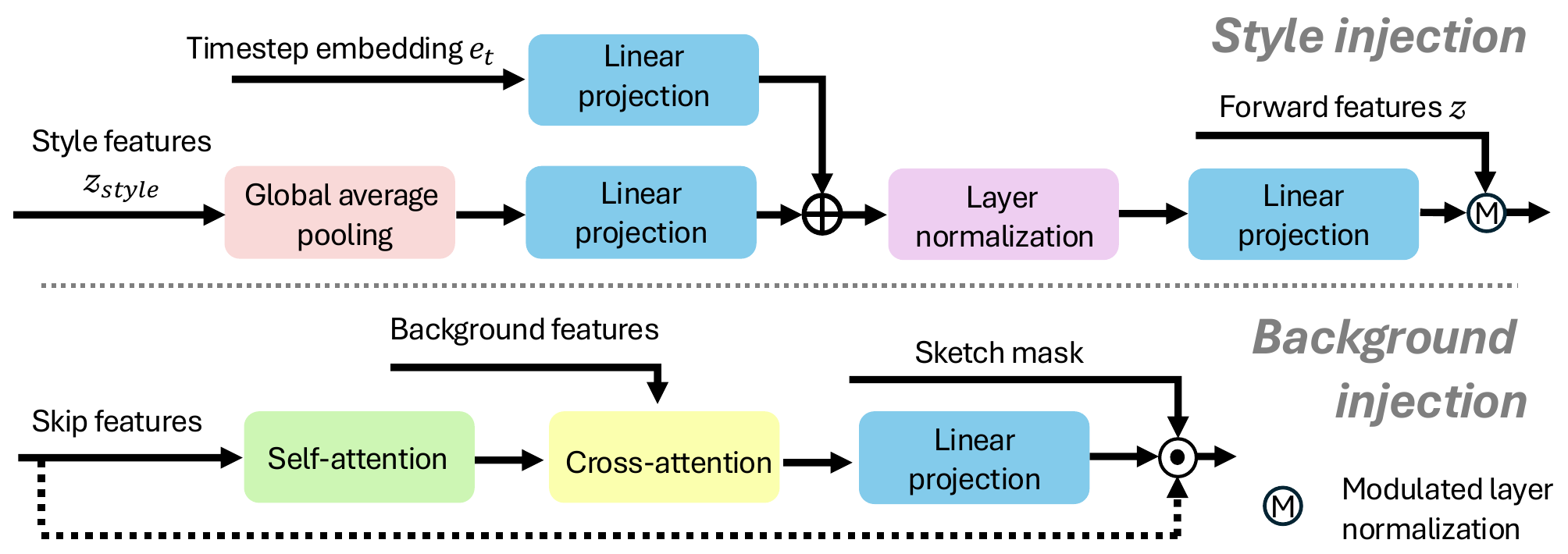}
    \caption{Illustration of background injection and style injection. The composition operation of style injection is expressed by Eq. \ref{warp-module}. For background injection, We first project the timestep embedding $\bm{e}_{t}$ and style features $\bm{z}_{style}$ into the same size, then perform element-wise addition and use the result for style modulation. The corresponding calculation is given in Eq. \ref{style_modulation-eq1} and \ref{style_modulation-eq2}.} 
    \label{modulation}
\end{figure}

To facilitate better feature injection and disentangle foreground and background features, thereby mitigating spatial entanglement, we further implement a background injection module. As illustrated in Figure \ref{framework}, the outputs of the background encoder at various latent levels are integrated with the skip connections of the denoising U-Net encoder via these background injection modules and subsequently concatenated at the corresponding levels of the denoising U-Net decoder. These background injection modules consist of a standard transformer block incorporating a spatial mask, as visualized in Figure \ref{modulation}. Denoting the computation of the transformer block as $\mathcal{W}(\cdot)$, the skip feature as $\bm{z}_{skip}$, and the feature from the background encoder as $\bm{z}_{bg}$. We formulate the calculation of the background injection modules as follows:
\begin{equation}
\begin{small}
    \bm{z}_{skip} = \begin{cases}
    \bm{z}_{skip} & \text{if $\bm{m_{s}} > ts_{s}$}\\
    \mathcal{W}(\bm{z}_{skip},\bm{z}_{bg}) & \text{if $\bm{m_{s}}\leq ts_{s}$}.
    \end{cases}
    \label{warp-module}
\end{small}
\end{equation}


\subsection{Style encoder and reference injection}
\label{style_encoder}
Aside from the background content, we also design a style injection module to integrate the extracted style features into the diffusion backbone to enhance global style transfer. After the diffusion backbone is pre-trained and the background encoder is merged into the framework, the forward features contain detailed backgrounds and a roughly colorized sketch foreground. We then train a style encoder to extract feature-level style information from reference images and then project the globally averaged latent features into the diffusion backbone, which is inferable from Table \ref{reference-comp}.

Adaptive normalization has been widely adopted to achieve better control in neural style transfer tasks \cite{GatysEB16,HuangB17,KarrasLA19,DiT}. Given the forward features in denoising U-Net as $\bm{z}$, the style features from the style encoder as $\bm{z}_{style}$, global average pooling as $\text{GAP}(\cdot)$, timestep embedding as $\bm{e}_{t}$, and layer normalization as $\mathcal{N}[\cdot]$, the style modulation $\mathcal{M}(\cdot)$ can be formulated as
\begin{equation}
    \mathcal{M}(\bm{z},\hat{\bm{z}}_{scale},\hat{\bm{z}}_{shift},\bm{e}_{t})=\bm{z}\cdot(1+\hat{\bm{z}}_{scale})+\hat{\bm{z}}_{shift},
    \label{style_modulation-eq1}
\end{equation}
where $\hat{\bm{z}}_{scale}$ and $\hat{\bm{z}}_{shift}$ calculated as
\begin{equation}
    \hat{\bm{z}}_{i}=W_{i}^{1}\mathcal{N}[W_{i}^{0}\text{GAP}(\bm{z}_{style})+(W^{e}_{i}\bm{e}_{t}+B_{i}^{e})]+B_{i}^{1},
    \label{style_modulation-eq2}
\end{equation}
and $i$ represents the index $scale$ or $shift$. Here, $W$ and $B$ are trainable parameters of the linear layers.

\begin{table*}[t]
    \centering
    \caption{Comparison of different inference modes.}
    \begin{tabular}{|p{2cm}|p{5.3cm}|p{4.3cm}|p{4cm}|}
        \hline 
        \centering Mode & \centering Reference carrier & \centering Advantage & \centering\arraybackslash Disadvantages\\
        \hline
        \centering\textit{Vanilla} & Full image: embeddings & Preserving sketch composition & Semantic spatial entanglement and details transfer \\
        \hline
        \centering\textit{Style enhance} & Full image: embeddings + feature-level modulation & Transferring style-related details, preserving sketch composition & Semantic spatial entanglement and background transfer \\
        \hline
        \centering\textit{Background enhance} & Foreground: embeddings \newline Background: embeddings + latent features & Eliminate semantic spatial entanglement, background transfer & Spatial entanglement for non-mask region, mask-based\\
        \hline
    \end{tabular}
    \label{inference-mode-table}
\end{table*}

\subsection{Inference mode}
\label{inference-mode-section}
To address a variety of use cases, the proposed framework provides three alternative inference schemes by activating different components: the \textit{Vanilla} mode, the \textit{Style enhance} mode, and the \textit{Background enhance} mode. We illustrate their differences in Table \ref{inference-mode-table} to briefly illustrate their differences.

\textit{Vanilla} mode is recommended for sketches with complicated compositions such as landscape, as it loyally recovers the semantic structure of the sketches. 
As the diffusion backbone in the framework employs image embeddings as reference carriers, it inherits the ability to colorize sketch images based on embedding-level reference information. This mode discards split cross-attention and additional encoders, so it suffers from spatial entanglement and is less effective in recovering details.

\textit{Style enhance} mode activates the style encoder during denoising and is able to transfer fine-grained textures and strokes. It is an upgraded version of \textit{Vanilla} mode, but it may degrade the semantic segmentation of results in certain corner cases.

\textit{Background enhance} mode is the mask-guided method. It activates the background encoder and the split cross-attention layers. As the foreground and background are explicitly separated in all forward passes in this mode, it effectively eliminates spatial entanglement and generates backgrounds with fine details and textures.

The two enhancement modes can be jointly activated as \textit{full enhance} to achieve the best transfer performance. Experimental results are given in Section 5 to compare their differences.

\begin{figure}[t]
    \centering
    \includegraphics[width=1\linewidth]{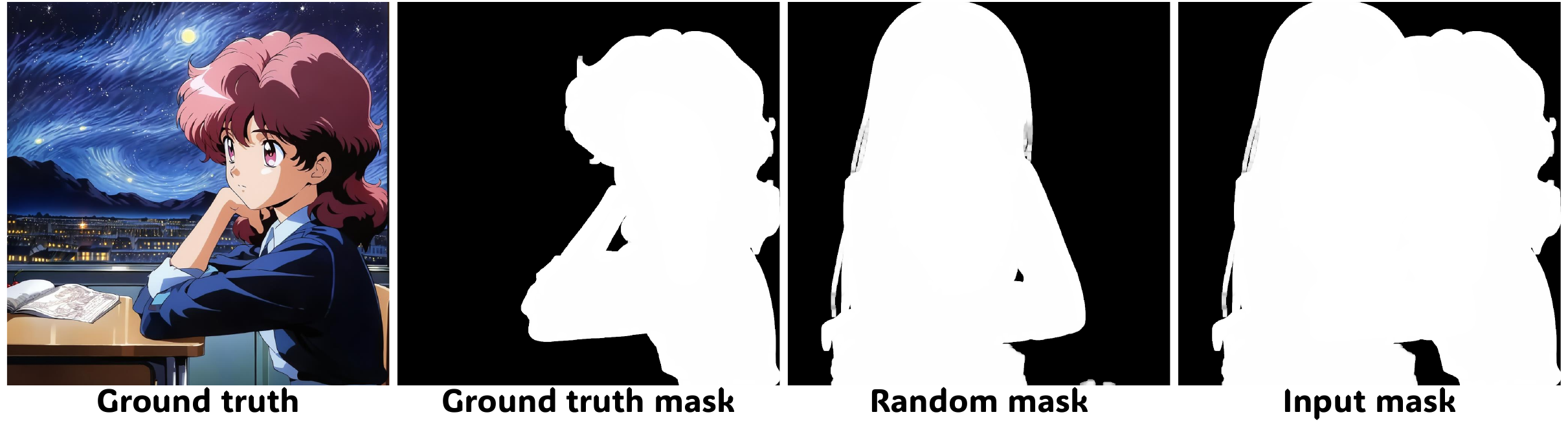}
    \caption{Visualization of the character-mask merging. We randomly merge the ground truth mask with another one as the input reference mask to optimize the networks for seamless foreground-background merging.}
    \label{chara-mask-merging}
\end{figure}
\begin{figure}[t]
    \centering
    \includegraphics[width=1\linewidth]{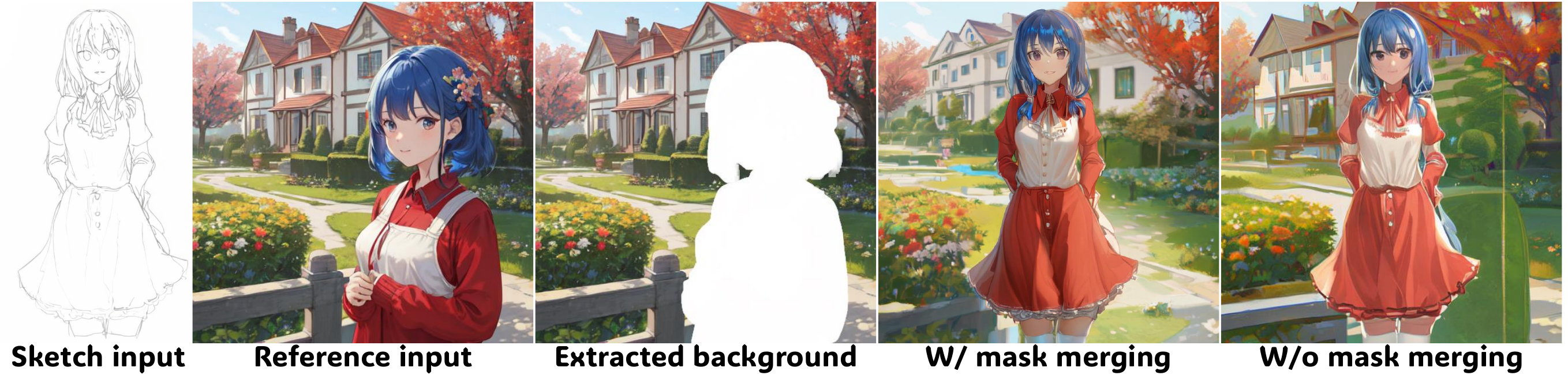}
    \caption{Comparison regarding mask merging. The ablation model trained without the mask merging cannot synthesize reasonable content in the extracted region.}
    \label{chara-mask-result}
\end{figure}

\begin{figure}[t]
    \centering
    \includegraphics[width=1\linewidth]{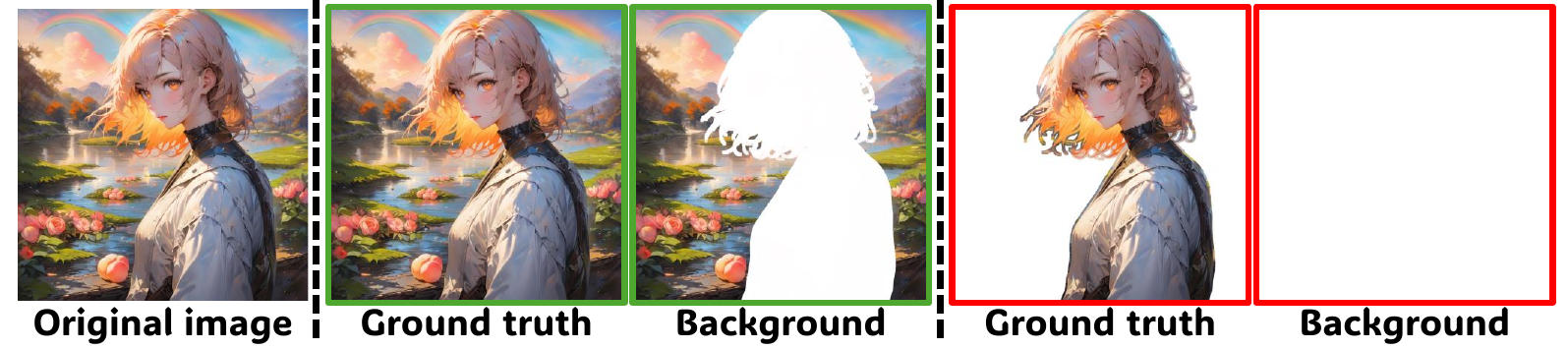}
    \caption{Visualization of the background bleaching. Green rectangle: original inputs; red rectangle: processed inputs.}
    \label{bleaching}
\end{figure}

\begin{figure}[t]
    \centering
    \includegraphics[width=1\linewidth]{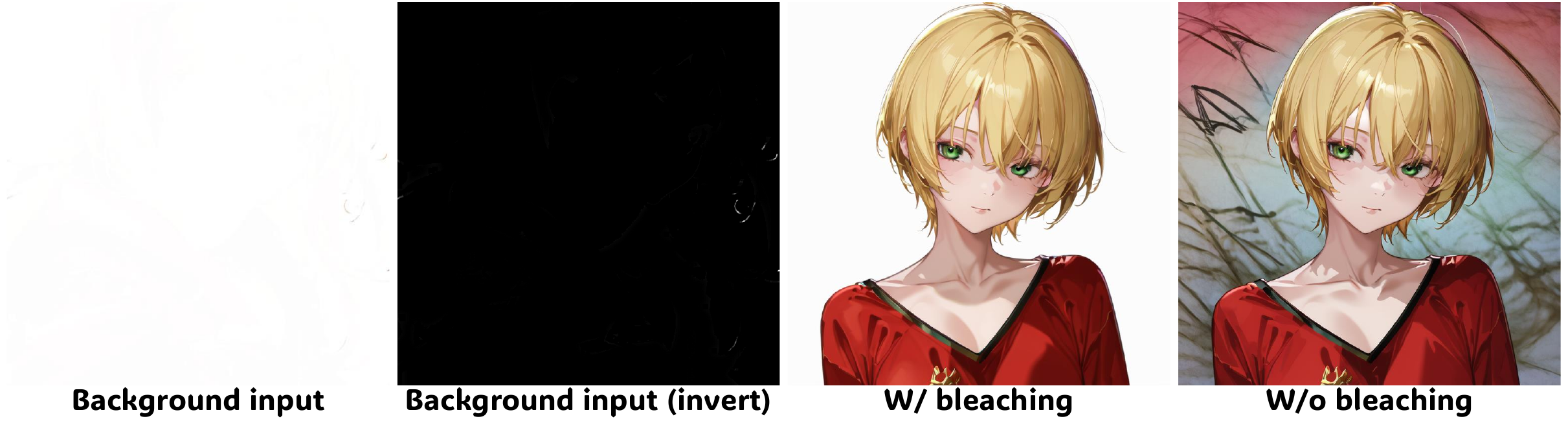}
    \caption{A comparison of results generated by networks trained with and without background bleaching, where the ablation framework failed to synthesize clean background with mostly blank reference. We invert the background input to increase the visibility of the trace.}
    \label{bg-bleach}
\end{figure}

\begin{figure}[t]
    \centering
    \includegraphics[width=1\linewidth]{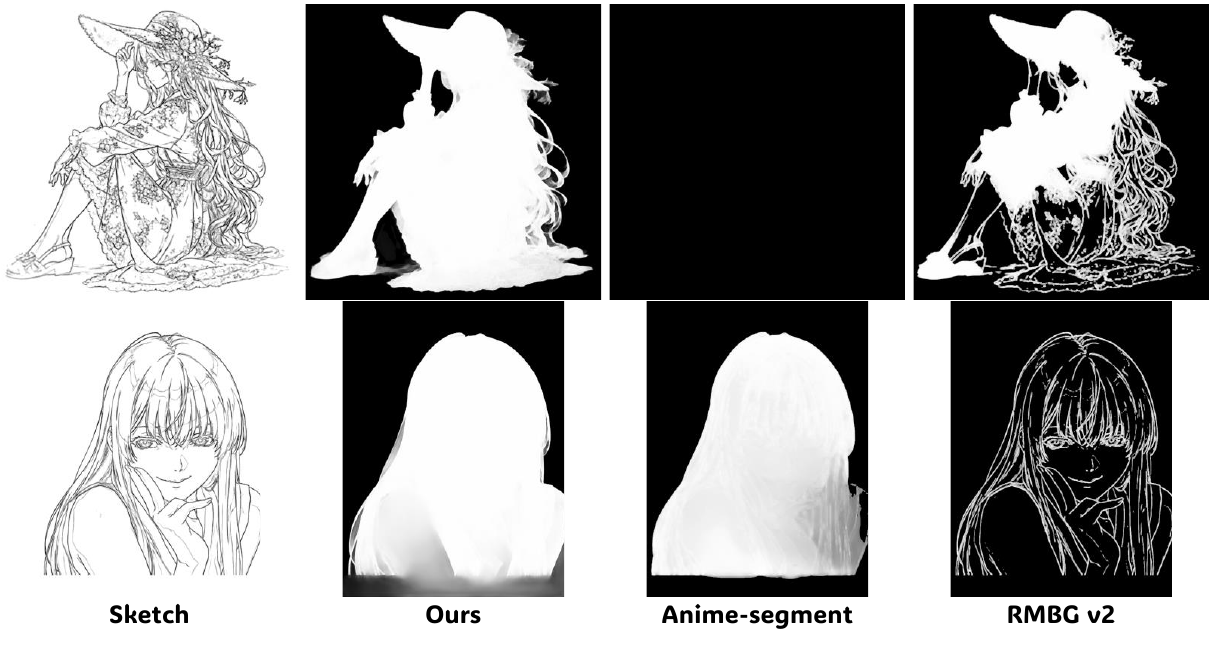}
    \caption{A comparison of sketch segmentation between our fine-tuned model with anime-segmentation \cite{isnet-eccv2022,anime-segmentation} and RMBG v2\cite{rmbg-BiRefNet}. Our segmentation network would extend the foreground for outpainting based on prior knowledge.}
    \label{sketch-segment}
\end{figure}

\begin{figure}[t]
    \centering
    \includegraphics[width=1\linewidth]{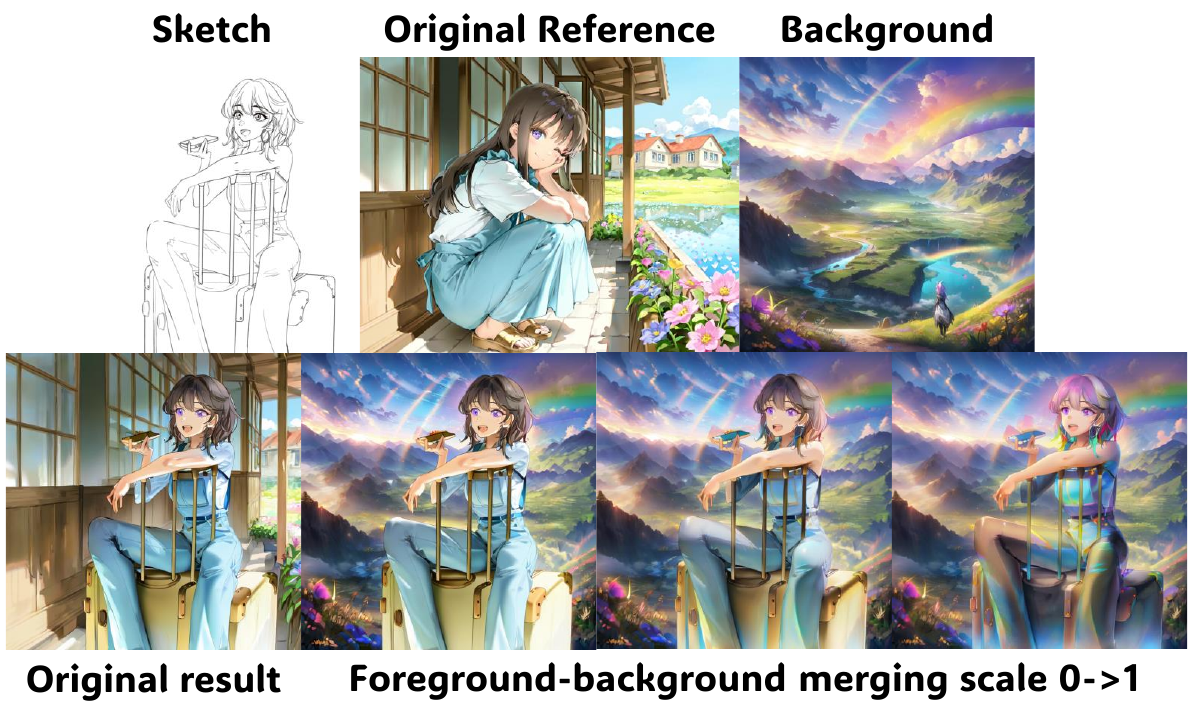}
    \caption{With the split cross-attention and background encoder, the proposed framework can use the other image as background reference and merge foreground-background with an adjustable scale.}
    \label{bg-replace}
\end{figure}

\section{Training strategy}
Compared to vanilla diffusion training, reference-based colorization requires sophisticated techniques to eliminate artifacts caused by spatial entanglement and realize clear foreground-background separation. In this section, we introduce the unique multi-stage strategy to train the framework and the pre-processing approach effective in further improving the colorization quality. 

\subsection{Step-by-step training}
The proposed framework consists of multiple components to realize corresponding functions. To better optimize each component, we propose to train them with a multiple-step training strategy: 
1. reference-based colorization pre-training, where only the sketch encoder and denoising U-Net are optimized. We follow the experience of \cite{yan2024colorizediffusion} to train the network with a noisy training-refinement training two-stage scheduler. We adopt a dynamic reference drop of 80\% for the noisy training stage and 50\% reference drop rate for refinement stage to avoid severe deterioration in the segmentation and perceptual quality of results 2. Foreground-background separation training for split cross-attention LoRAs and the background encoder, which helps eliminate spatial entanglement caused by the reference embeddings and enhance the recovery of backgrounds; 3. Hybrid training for the style encoder, where the background encoder and split cross-attention are not optimized but randomly activated at a rate of 50\% to generate extra conditions for the denoising backbone. In stage 2 and stage 3, the reference embeddings for denoising U-Net was dropped at a fixed rate of 50\%. These two steps play important roles in enhancing the colorization performance in the proposed framework.

Given noise $\epsilon$, sketch $\bm{s}$, ground truth $\bm{y}$, encoded latent representations (forward features) $\bm{z}_{t}$ at timestep $t$, VAE encoder $\mathcal{E}$, denoising U-Net $\theta$, background encoder with background injection $\varphi_{bg}$, style encoder with style injection $\varphi_{style}$, and CLIP image encoder $\phi$, the training objective for all training stages can be defined as 
\begin{equation}
    \mathbb{E}_{\mathcal{E}(\bm{y}),\epsilon,t,\bm{s},\bm{c}}[\|\epsilon-\epsilon_{\theta}(\bm{z}_{t},t,\bm{s},\bm{c})\|^{2}_{2}],
    \label{diffusion-loss}
\end{equation}
where $\bm{c}$ represents the accumulated conditional inputs. Specifically, the conditional inputs are composed of the following terms at each stage:

\begin{itemize}[leftmargin=*, label={}]
    \item \textbf{Stage 1}: image embeddings $\bm{e}$.
    \item \textbf{Stage 2}: background embeddings $\bm{e}_{bg}$, sketch mask $\bm{m}_{s}$, and background features $\bm{z}_{bg}=\varphi_{bg}\big(\mathcal{E}(\bm{r}_{bg}),\bm{e}\big)$.
    \item \textbf{Stage 3}: style features $z_{style}=\varphi_{style}\big(\mathcal{E}(\bm{r}),\bm{e}\big)$.
\end{itemize}
The training strategy is carefully designed to strictly limit the spatial semantics of style features $z_{style}$, thereby mitigating spatial entanglement. Note that in our implementation, we directly utilized ground truth as reference inputs during training, without any augmentation.

\subsection{Preprocessing of conditional inputs}
The background inputs to train the background encoder are images with foreground regions removed. However, simply erasing foreground regions from input images is observed to hinder the encoder's ability to seamlessly merge foreground and background or generate satisfactory inpainting results. Consequently, we propose \textbf{character-mask merging} to address this issue. As visualized in Figure \ref{chara-mask-merging}, it randomly adds a mismatched character mask on each ground truth image.

This preprocessing strategy compels the background encoder to perform inpainting and the background injection modules to facilitate the seamless integration of foreground elements with the character-extracted backgrounds. A comparison of this preprocessing technique is presented in Figure \ref{chara-mask-result}.

Furthermore, we address a specific use case where users aim to colorize only the foreground by providing reference images with clear backgrounds and sketches devoid of background lines. We observed that image-guided networks, particularly when trained with the proposed \textbf{chracter-mask mergin} preprocessing, exhibit a tendency to synthesize noisy backgrounds, especially when utilizing predominantly blank reference images. 

To mitigate this limitation, we further propose \textbf{background bleaching} preprocessing technique which is visualized in Figure \ref{bleaching}. By randomly bleaching the background of ground truth and reference inputs during training at a rate of 1\%, it effectively eliminates the artifacts on the background for similar use cases. A comparison is given in Figure \ref{bg-bleach} to demonstrate the importance of this preprocess.

\begin{figure}[t]
    \centering
    \includegraphics[width=1\linewidth]{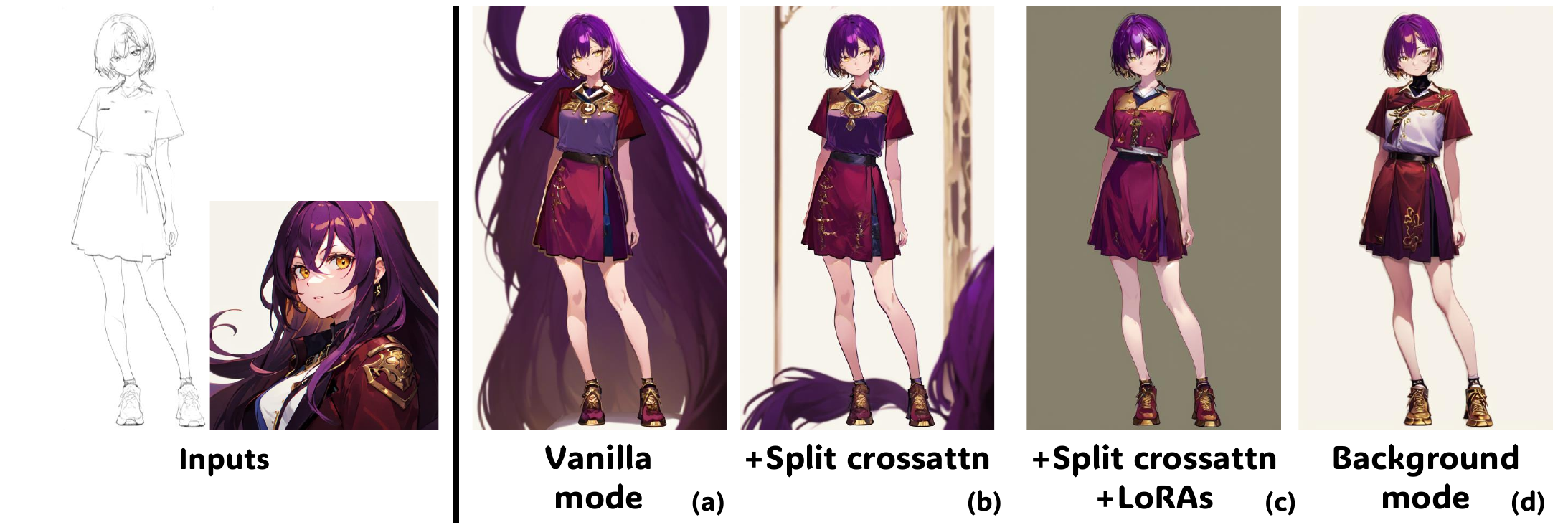}
    \caption{Spatial entanglement artifacts from image embeddings can be eliminated by split cross-attention and foreground-background LoRAs. We further improve the background transfer using the background encoder.}
    \label{ablation}
\end{figure}
\begin{figure}[t]
    \centering
    \includegraphics[width=1\linewidth]{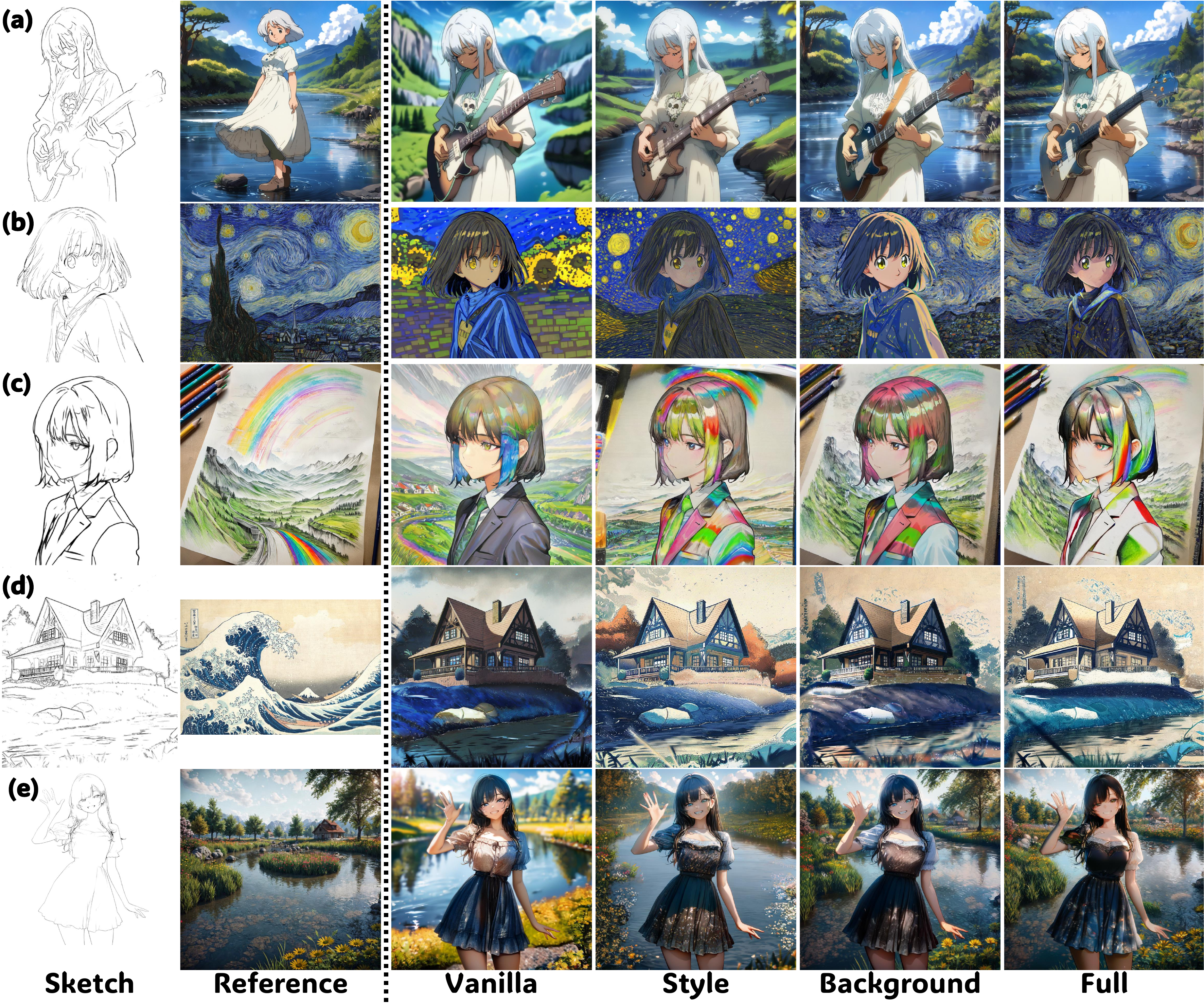}
    \caption{Colorization results with different inference modes. Both \textit{style enhance} and \textit{background enhance} modes transfer feature-level information, enabling the framework to synthesize more vivid textures. High-resolution images are available in the supplementary materials for checking style details.}
    \label{inference-mode}
\end{figure}
\begin{figure*}[t]
    \centering
    \includegraphics[width=1\linewidth]{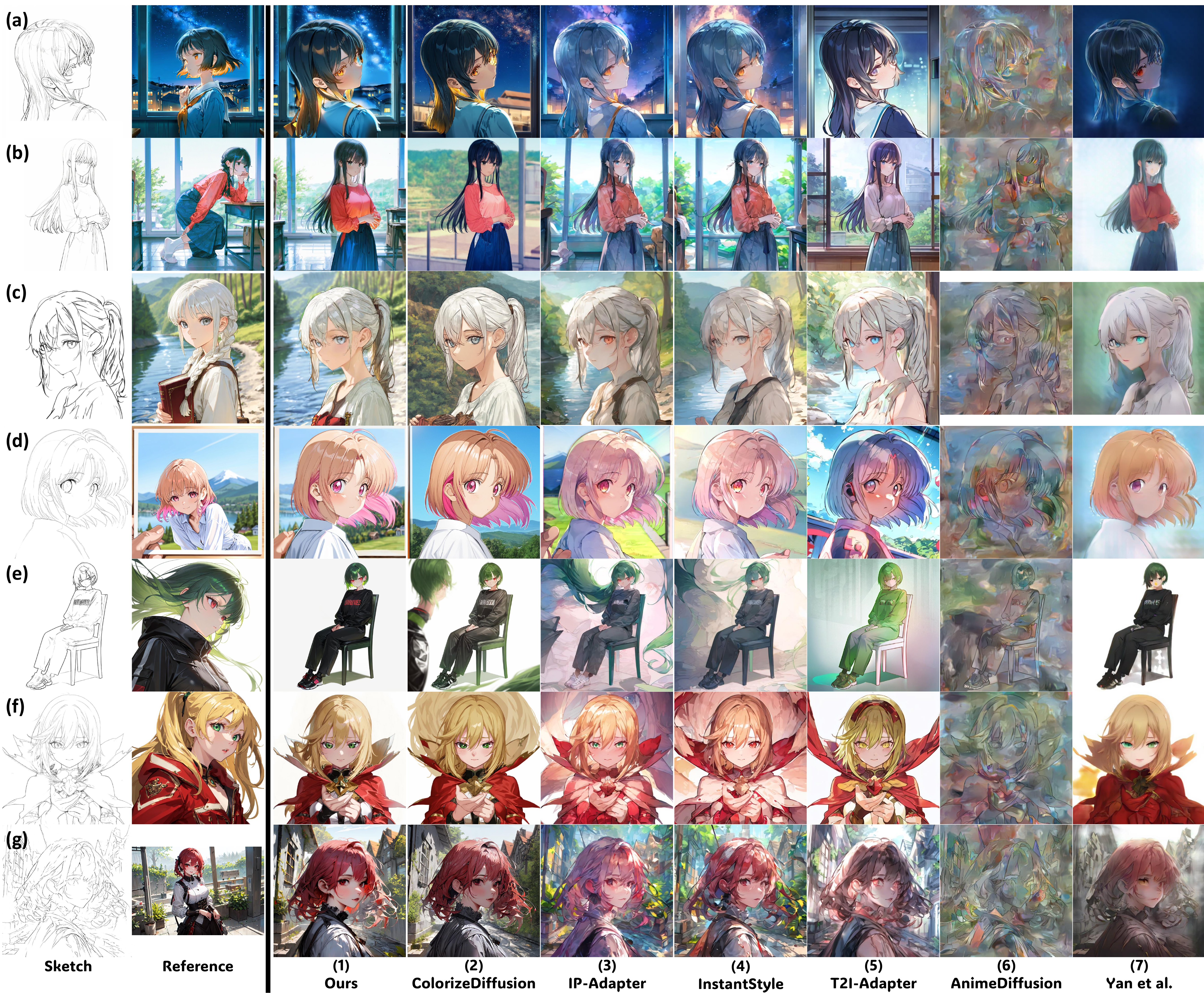}
    \caption{Qualitative comparisons regarding figure colorization. Our results demonstrate the superiority of the proposed framework in improving the quality and similarity of colorization without having spatial entanglement, which can be clearly observed in rows (e)-(f). Zoom in for details.}
    \label{qualitative}
\end{figure*}

\section{Experiment}
\subsection{Implementation}
\noindent\textbf{Dataset, configurations, and environment.} We curated a dataset of over 6M (sketch, color, mask) image triples from Danbooru \cite{danbooru2021}, encompassing a diverse range of anime styles. Sketches were generated by jointly using \cite{sketchKeras} and \cite{xiang2022adversarial}, while masks were produced using \cite{anime-segmentation}. The dataset was divided into a training set and a validation set of 51,000 triples without overlap data.

Our training was performed on 8x H100 (94GB) GPUs utilizing Deepspeed ZeRO2 \cite{deepspeed} and the AdamW optimizer \cite{KingmaB14,LoshchilovH19} with a batch size of 256, a learning rate of 0.00001, and betas of (0.9, 0.999). All training inputs were initially resized to 800$\times$800 and then randomly cropped to 768$\times$768, with the exception of reference images, which were resized directly.\\

\noindent\textbf{Fine-tuning of sketch segmentation network.} Existing background removal methods cannot accurately extract the boundaries of sketches. To address this limitation, we fine-tuned a segmentation network \cite{anime-segmentation,isnet-eccv2022} using our (sketch, mask) pairs. This fine-tuning enables precise foreground-background separation for sketch images and allows adaptive extension of the foreground boundary for sketch-based outpainting. A visual comparison is provided in Figure \ref{sketch-segment}. Given the significance of sketch segmentation for the proposed framework, we will also release the fine-tuned weights.\\

\noindent\textbf{Background replacement and merging.} Our framework facilitates flexible background replacement and style merging through independent processing of foreground and background. Users can perform seamless replacement of the original background with a different reference while simultaneously merging styles using a controllable scale. Figure \ref{bg-replace} illustrates this process. More details are included in the supplementary materials.

\begin{figure}[t]
    \centering
    \includegraphics[width=1\linewidth]{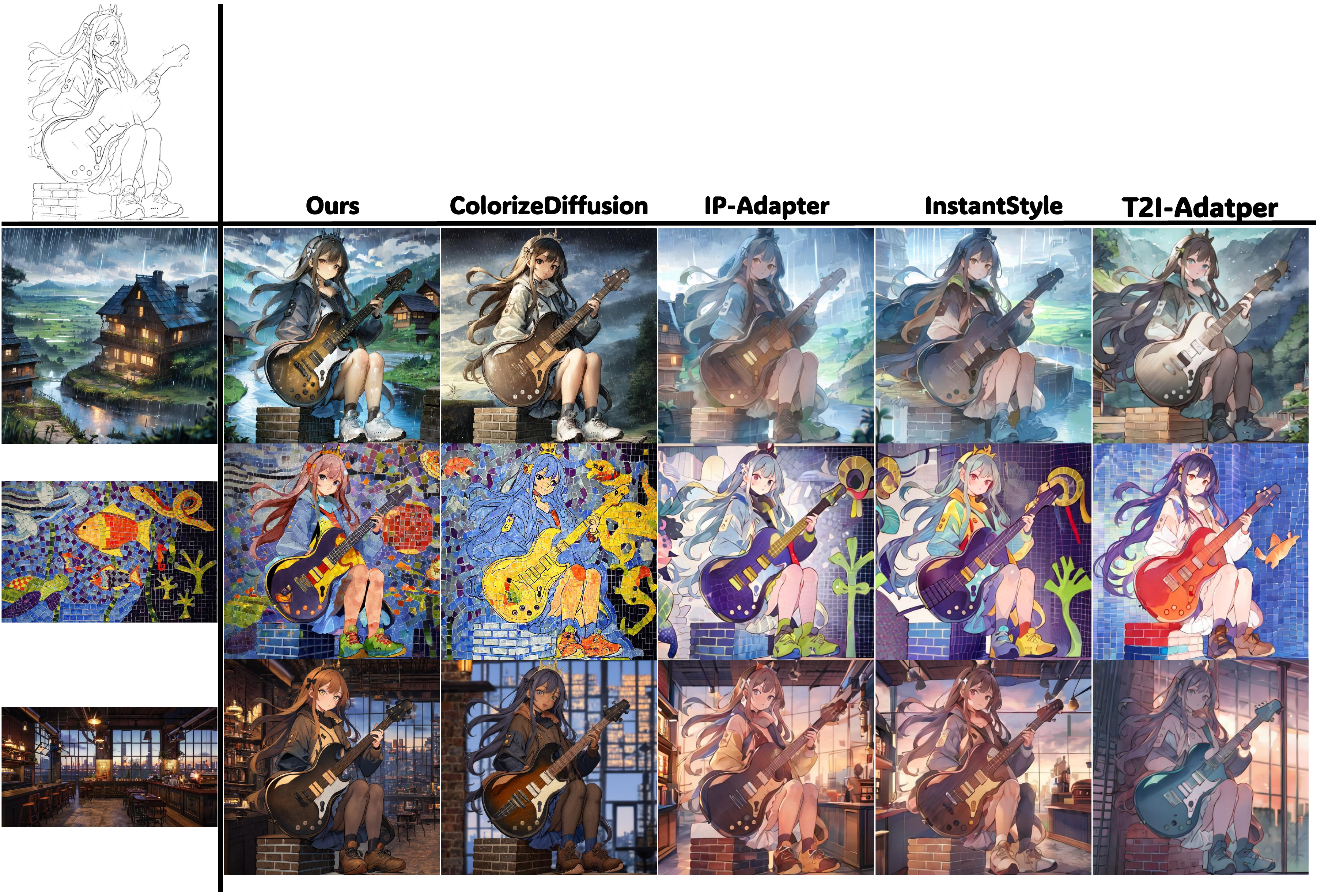}
    \caption{Cross-content colorization for figure sketch, our results significantly outperform baseline methods in character segmentation and colorization.}
    \label{cross-qualitative-1}
\end{figure}

\subsection{Ablation study}
\noindent\textbf{Artifacts removal.} We examine the necessity of the split cross-attention and background encoder for \textit{background enhance} mode, whose first target is to eliminate artifacts caused by the spatial entanglement from local embeddings \cite{yan2024colorizediffusion}. We set up four ablation frameworks: 1) \textit{vanilla} mode, 2) \textit{vanilla} mode with split cross-attention but no trainable LoRAs and background encoder, 3) \textit{vanilla} mode with split cross-attention and trainable LoRAs, and 4) \textit{background enhance} mode.

We show a qualitative comparison in Figure \ref{ablation} to validate the effectiveness of the proposed modules. The ablation models generated additional hair in (a) and (b). While adopting foreground-background LoRAs eliminates the spatial entanglement, it falls short in transferring backgrounds compared to the proposed \textit{background enhance} mode due to the absence of the background encoder, which transfers more detailed background features.\\

\noindent\textbf{Inference modes.} Figure \ref{inference-mode} illustrates the distinctions between inference modes. In \textit{vanilla} mode, colorization relies solely on local image embeddings from the CLIP image encoder. This mode resembles T2I generation, primarily retrieving color attributes from the training data rather than transferring them from reference images. Consequently, \textit{vanilla} mode struggles to replicate intricate stroke details from references that deviate from the anime style, as observed in rows (b), (d), and (e). In contrast, \textit{style enhance} and \textit{background enhance} modes utilize latent features as carriers, enabling them to transfer more detailed information from references for global style and background, respectively.

As introduced in Section \ref{inference-mode-section}, \textit{style enhance} and \textit{background enhance} can be jointly used to generate detailed backgrounds and vivid textures globally.

\begin{figure}[t]
    \centering
    \includegraphics[width=1\linewidth]{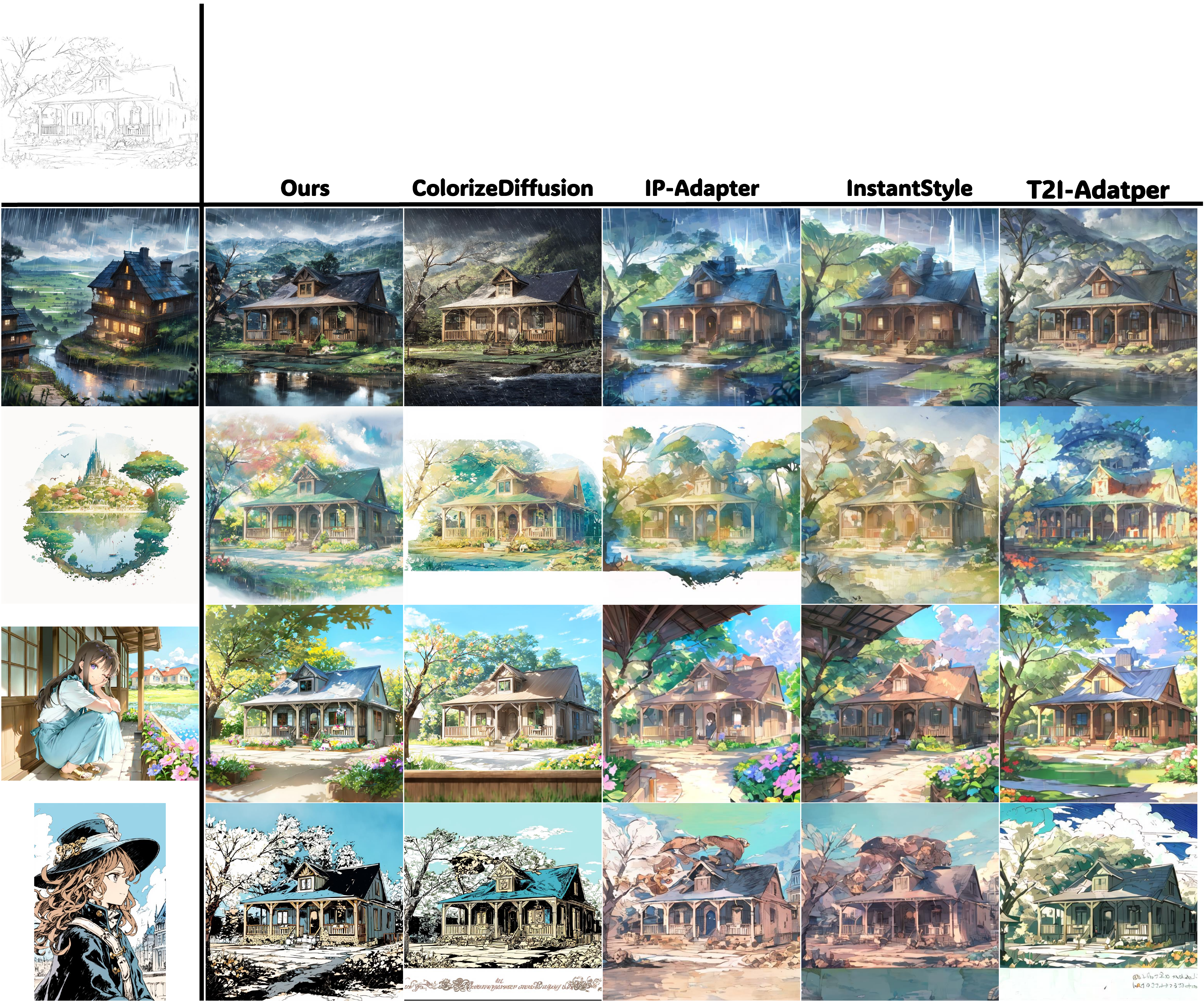}
    \caption{Cross-content colorization for landscape sketch. Our results demonstrate better composition and style transfer performance, as well as outpainting for non-sketch regions.}
    \label{cross-qualitative-2}
\end{figure}

\begin{table*}[t]
    \centering
    \caption{Quantitative comparison between the proposed model and baseline methods. We calculated 50K aesthetic score \cite{aesthetic-predictor}, 50K FID \cite{Seitzer2020FID}, 5K PSNR, 5K MS-SSIM, and 5K CLIP cosine similarity of image embeddings in this experiment. \dag: These evaluations randomly selected color images as references, making them close to real-application scenarios. \ddag: Ground truth color images were deformed to obtain semantically paired and spatially similar references for evaluations.}
    \begin{tabular}{|c|c|c|c|c|c|c|c|c|c|}
        \hline
        \multicolumn{5}{|c|}{Method} & {\dag Aesthetic score $\uparrow$} & {\dag FID $\downarrow$} & {\ddag PSNR$\uparrow$} & {\ddag MS-SSIM$\uparrow$} & {\ddag CLIP similarity$\uparrow$}\\
        \hline
        \multicolumn{5}{|c|}{Ours - \textit{full enhance}} & \textbf{5.1859} &\textbf{5.6330} & \textbf{29.3626} & \textbf{0.7081} & \textbf{0.9056} \\
        \hline
        \multicolumn{5}{|c|}{\textit{ColorizeDiffusion}} & 4.8351 & 9.6423 & 28.7215 & 0.5899 & 0.8753 \\
        \hline
	\multicolumn{5}{|c|}{\textit{IP-Adapter}} & 4.6627 & 38.9232 & 28.5124 & 0.5464 & 0.8632 \\
        \hline
        \multicolumn{5}{|c|}{\textit{InstantStyle}} & 4.7150 & 40.2134 & 28.0921 & 0.4467 & 0.8039 \\
	\hline
        \multicolumn{5}{|c|}{\textit{T2I-Adapter}} & 4.2647 &41.1569 & 28.1321 & 0.3194 & 0.7134 \\
        \hline
        \multicolumn{5}{|c|}{\textit{AnimeDiffusion}} & 4.2626 &63.4123 & 27.7459 & 0.3183 & 0.7304 \\
	\hline
        \multicolumn{5}{|c|}{Yan et al.} & 4.7923 &27.0032 & 29.1293 & 0.5239 & 0.7599\\
	\hline
	\end{tabular}
    \label{quantitative}
\end{table*}

\subsection{Comparison with baseline}
We compare our method with existing reference-based sketch image colorization methods \cite{yan-cgf,animediffusion,yan2024colorizediffusion,ip-adapter,controlnet-iccv,t2i-adapter} to demonstrate the superiority of the proposed framework. Given the complexity of adapter-based methods, we provide a concise overview of their operation and integration into the image-guided sketch colorization before our experimental analysis.\\

\noindent\textbf{Adapter-based baseline.} Reference-based sketch colorization can be achieved by combining Stable Diffusion (SD) \cite{RombachBLEO22}, ControlNet \cite{controlnet-iccv,controlnet-v11}, and IP-Adapter/T2I-Adapter \cite{ip-adapter,t2i-adapter}, where ControlNet provides sketch control and IP-Adapter/T2I-Adapter extract visual features from image prompts for color, style, and identity reference. Based on their prompt adapter, we label these baseline methods as \textit{IP-Adapter} and \textit{T2I-adapter}. InstantStyle \cite{instantstyle} suggests that adjusting cross-attention scales of \textit{IP-Adapter} may help diminish spatial entanglement by reducing the transfer of composition information. We therefore include \textit{InstantStyle} as a baseline to validate its effectiveness. We select \textit{Anything v3} as the SD backbone for \textit{IP-Adatper}, \textit{InstantStyle}, and \textit{T2I-Adapter} since the anime-style sketch ControlNet is trained with it \cite{controlnet-v11}. We only consider SD 1.5-based variations due to the lack of more effective combinations for other diffusion backbones. 

The IP-Adatper-based methods utilize latent features and image embeddings as carriers jointly since it is initially optimized for image reconstruction, where its trainable modules, therefore, recover feature-level information after extracting embeddings from the CLIP image encoder.\\

\noindent\textbf{Qualitative comparison.} We present a qualitative comparison of figure colorization results in Figure \ref{qualitative}. Rows (a)-(d) demonstrate the superior performance of our proposed framework across all dimensions, including background reconstruction, accurate color representation, and effective style transfer. The removal of artifacts arising from spatial entanglement is evident in rows (e)-(f) when comparing our results with those of ColorizeDiffusion, IP-Adapter, InstantStyle, and T2I-Adapter. We further highlight a pair of inputs with complex backgrounds in the sketch in row (g). Among the baselines, AnimeDiffusion \cite{animediffusion} is specifically designed for face colorization. Its training process, which concatenates references with sketches, easily leads to overfitting on training sets with restricted content, hindering the generation of diverse images. The GAN-based method \cite{yan-cgf} struggles to synthesize accurate colors and backgrounds for complex inputs due to limitations in its neural backbone. Compared to these baselines, our framework consistently generates high-quality colorized images with strong similarity to the reference images while effectively avoiding artifacts across a wide range of content.

We continue to exhibit cross-content colorization of figure and landscape sketches in Figures \ref{cross-qualitative-1} and \ref{cross-qualitative-2}. In these scenarios, subjective evaluations usually prioritize the similarity of style and color scheme due to the limited correspondence between input identities. The proposed framework significantly outperforms baseline methods in this regard while closely adhering to the sketch semantics to achieve clearer visual segmentation outcomes.\\

\noindent\textbf{Quantitative comparison.} Aesthetic score \cite{aesthetic-predictor} and Fréchet Inception Distance (FID) \cite{HeuselRUNH17} are widely used quantitative metrics to evaluate the perceptual quality of synthesized results without requiring inputs to be semantically and spatially paired. We conducted two evaluations using these metrics on the entire validation set, which contains 52k+ (sketch, reference) image pairs. Reference images were randomly selected during the validation.

Besides, we also tested multi-scale structural similarity index measure (MS-SSIM), peak signal-to-noise ratio (PSNR), and CLIP score \cite{RadfordKHRGASAM21,openclip} for assessing the similarity between generated images and given ground truth. As these metrics require the reference to be aligned with ground truth, we selected 5000 color images as ground truth to generate extracted sketches and deformed references, where references were deformed using thin plate spline (TPS) transformation.

We conclude the quantitative evaluation results in Table \ref{quantitative}, where the proposed method significantly outperforms in all evaluations owing to the removal of artifacts, higher fidelity to the sketch composition, and stronger style transfer ability.\\

\begin{figure}[t]
    \centering
    \includegraphics[width=\linewidth]{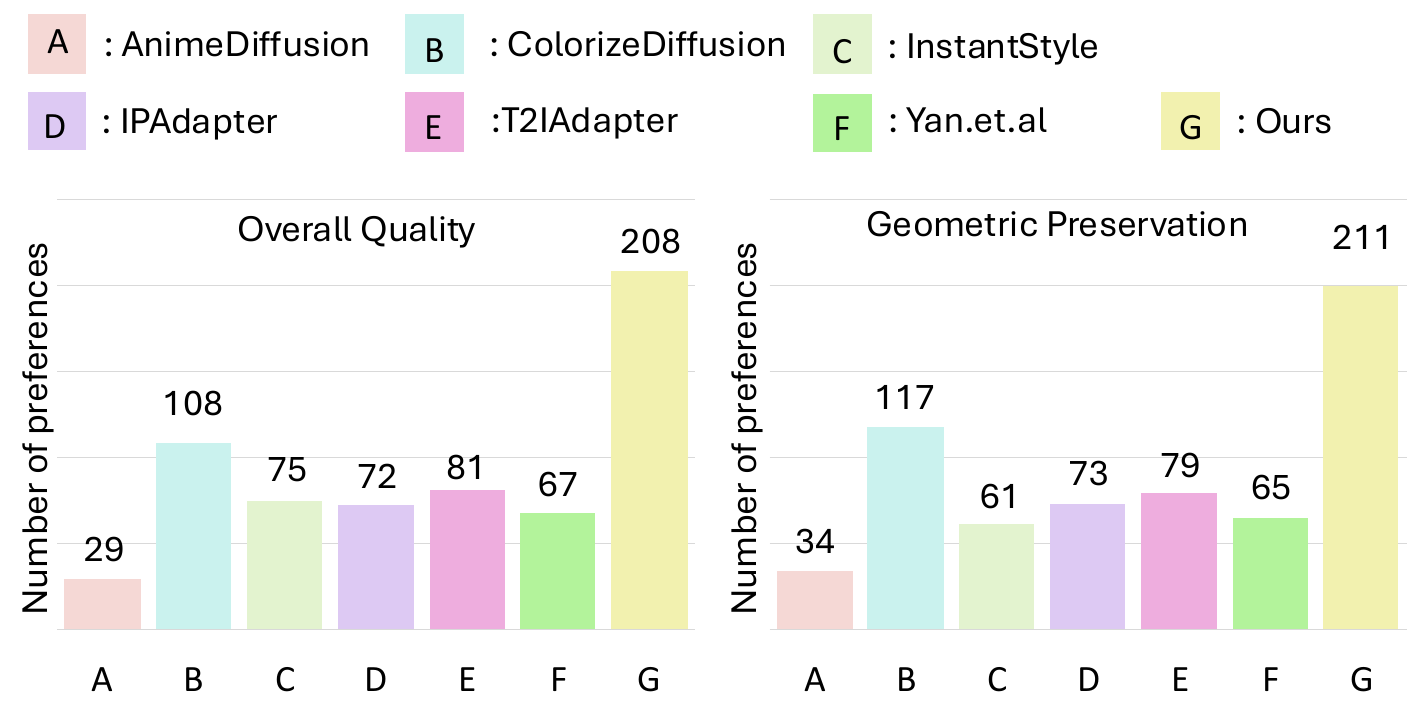}
    \caption{Results of user study. Our method is preferred across all shown methods in overall quality and geometric preservation.}
    \label{userstudy}
\end{figure}

\noindent\textbf{User study.} To further reveal the subjective evaluation of the proposed method and existing methods by real persons, we demonstrate a user study with 40 participants invited to select the best results with two criteria: the overall colorization quality and the preservation of the geometric structure of the sketches. 25 image sets are prepared and each participant is shown 16 image sets for evaluation. We present to participants the colorization results of the proposed method and those generated by six existing methods for each image set.

We present the results of the user study in Figure \ref{userstudy}, with the results showing that our proposed method has received the most numbers of preferences across all the methods illustrated. For further validation of the comparison, the Kruskal-Wallis test is employed as a statistical method. The results demonstrate that our proposed method outperforms all previous methods significantly in terms of user preference with a significance level of p \textless 0.01. All the images shown in the user study are included in the supplementary materials.

\begin{figure}[t]
    \centering
    \includegraphics[width=1\linewidth]{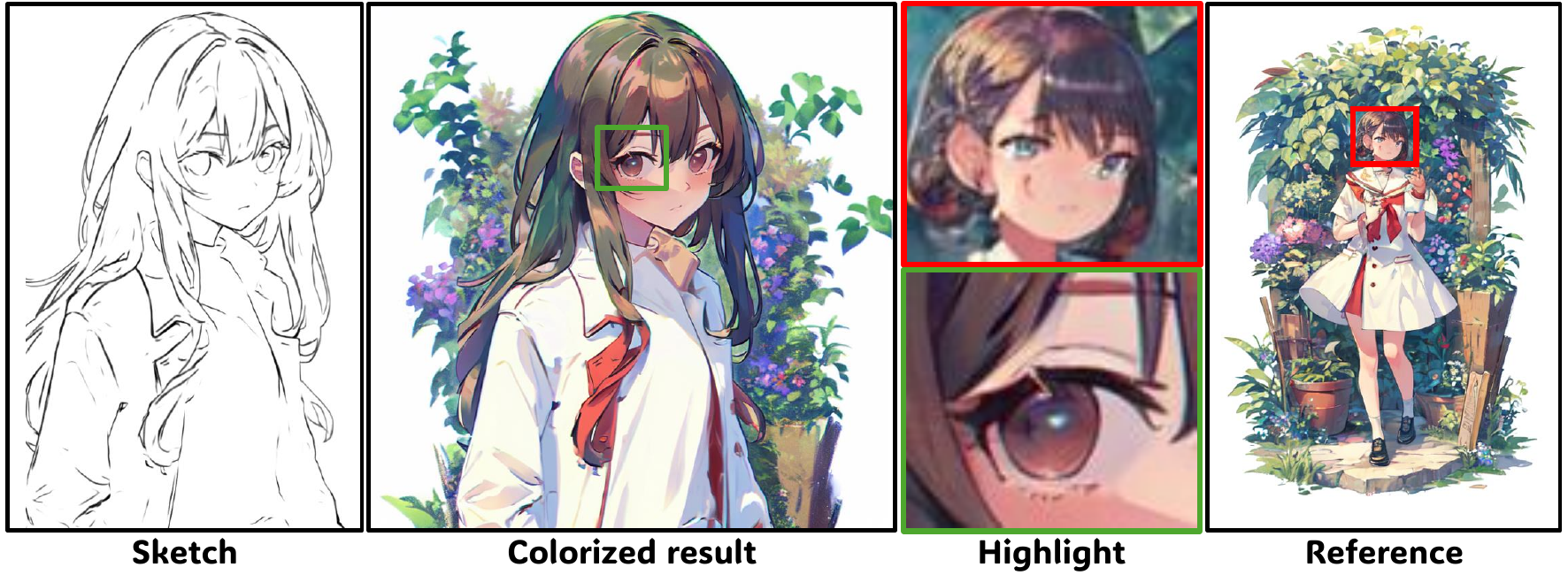}
    \caption{A failure case of eye colorization. Due to the small size of visual attributes in the reference image, the employed CLIP image encoder, with a 224x224 input size and a general-purpose design, struggles to extract relevant embeddings from anime-style images, leading to a suboptimal similarity.}
    \label{failure}
\end{figure}

\section{Conclusion}
This paper presents an image-guided sketch colorization framework designed to achieve state-of-the-art results with arbitrary inputs. To address the limitations of existing methods, we introduce split cross-attention to solve spatial dependencies and employ two independent encoders: one for background transfer and another for style modulation. This architecture enables zero-shot colorization for references in any style, even for realistic images.

A key limitation of the proposed framework lies in achieving high-precision character colorization. Since character colors are primarily determined by embeddings extracted from the CLIP image encoder, discrepancies may arise between the generated results and the reference image, as illustrated in Figure \ref{failure}. Our future work will focus on improving identity consistency in character colorization and text-based manipulation. Furthermore, we aim to extend this framework to enable robust video colorization.

{\small
\bibliographystyle{ACM-Reference-Format} 
\bibliography{egbib}
}

\end{document}